\documentclass[pdflatex,sn-mathphys-ay]{sn-jnl}

\usepackage{graphicx}%
\usepackage{multirow}%
\usepackage{amsmath,amssymb,amsfonts}%
\usepackage{amsthm}%
\usepackage{mathrsfs}%
\usepackage[title]{appendix}%
\usepackage{xcolor}%
\usepackage{textcomp}%
\usepackage{manyfoot}%
\usepackage{booktabs}%
\usepackage{algorithm}%
\usepackage{algorithmicx}%
\usepackage{algpseudocode}%
\usepackage{listings}%
\usepackage{hyperref}
\usepackage{appendix}
\usepackage{array}
\usepackage{caption}
\usepackage{xcolor}   

\begin{document}

\title{Unsupervised machine learning for data-driven rock mass classification: addressing limitations in existing systems using drilling data}

\author*[1,2]{\fnm{Tom F.} \sur{Hansen}}\email{tom.frode.hansen@ngi.no}

\author[1]{Arnstein Aarset}

\affil*[1]{\orgdiv{Engineering geology and rock engineering}, \orgname{Norwegian Geotechnical Institute}, \orgaddress{\street{Sandakerveien 140}, \city{Oslo}, \postcode{0484}, \country{Norway}}}

\affil[2]{\orgdiv{Informatics Institute}, \orgname{University of Oslo}, \orgaddress{\street{Blindern}, \city{Oslo}, \postcode{0316},\country{Norway}}}

\abstract{Rock mass classification systems are crucial for assessing stability and risk in underground construction globally and guiding support and excavation design. However, these systems, developed primarily in the 1970s, lack access to modern high-resolution data and advanced statistical techniques, limiting their effectiveness as decision-support systems. We outline these limitations and describe how a data-driven system, based on drilling data, can overcome them. Using statistical information extracted from thousands of MWD-data values in one-meter sections of a tunnel profile, acting as a signature of the rock mass, we demonstrate that well-defined clusters can form a foundational basis for various classification systems. Representation learning was used to reduce the dimensionality of 48-value vectors via a nonlinear manifold learning technique (UMAP) and linear principal component analysis (PCA) to enhance clustering. Unsupervised machine learning methods (HDBSCAN, Agglomerative Clustering, K-means) clustered the data, with hyperparameters optimised through multi-objective Bayesian optimisation. Domain knowledge improved clustering by adding extra features to core MWD-data clusters. We structured and correlated these clusters with physical rock properties, including rock type and quality, and analysed cumulative distributions of key MWD-parameters to determine if clusters meaningfully differentiate rock masses. The ability of MWD data to form distinct rock mass clusters suggests substantial potential for future classification systems using this objective, data-driven methodology, minimising human bias.}


\maketitle

\noindent\textbf{Highlights}

\begin{itemize}
    \item Natural clustering in Measure While Drilling data is demonstrated.
    \item Unsupervised clustering is optimised with multi-objective Bayesian optimisation.
    \item Clustering is sensitive to feature sets and algorithms for dimension reduction and clustering.
    \item Clusters are investigated and organised by physical features.
    \item A system approach is sketched for clusters as a foundational basis for rock mass classification systems.
\end{itemize}

\section{Introduction\label{introduction}}

Rock mass classification systems (RMCS) are widely used as decision support in rock tunnelling. The Q-system \citep{Barton1974} and the RMR \citep{Bieniawski1973} system are among the most popular \citep{Erharter2023RockMass}. In Scandinavian drill and blast tunnelling, operations typically involve full profile blasting for 5 to 6 meters, followed by 1-2 hours of scaling to remove loose rock. Subsequently, the surface is washed and coated with fibre-reinforced shotcrete roughly an hour later. Once the rock is covered, further physical inspection is not possible. Within this brief period, the face engineer must inspect the rock surface, evaluate input variables (considering a range to accommodate uncertainty), compute a classification value, and assign a specific stability class. This classification directly influences the support class, which dictates the overall rock support strategy. Temporary and permanent supports are usually installed simultaneously, underscoring the need for precise rock mass classification to maintain tunnelling efficiency and stability. The focus is primarily on the newly exposed rock contour, where support systems are implemented, typically an area of $150\,m^2$ (calculated as $25\,m \text{ arclength} \times 6\,m$). Less emphasis is placed on the tunnel face. However, the focus on the face increases in weaker rock masses where shorter lengths are standard, and advance support becomes critical. Although this description is specific to Scandinavian contexts, the findings and methodologies are equally applicable to other rock tunnelling regimes.

\subsection{Limitations in existing rock mass classification systems} \label{limitations}

RMCSs have been essential for the rock engineering industry and have contributed to increased consistency in rock mass assessment and rock support worldwide for decades. Developed mainly in the 1970s, these systems predate modern data capture technologies such as comprehensive scan/image capturing of newly exposed rock surface profiles, Measure While Drilling (MWD) data, and geophysics from the excavation face. At that time, high-resolution datasets with extensive rock mass coverage and advanced statistical learning techniques were unavailable, and computational power was limited. Given the current advances in data availability and automation potential in Scandinavian tunnelling, these classification systems exhibit limitations in design, and in how the systems are used, that may result in suboptimal decisions. Modern data-driven approaches could likely address these limitations. The objective is to highlight the enhancements and new approaches possible with today's technology, available data and computational power, not to undermine the existing systems. The following listed limitations must be seen in that context. 

\begin{enumerate}
    \item Subjectivity in Assessment: These systems rely heavily on the subjective judgement of face engineers, leading to variations in the assessment and support decisions for identical rock conditions \citep{Erlend2023subjective,SEN2003269, Elmo2021, ambah_is_2024, stille_classification_2003}.\label{item:human_bias}
    \item Inconsistent Observations: Face engineers may focus inconsistently on specific features or areas, potentially overlooking variations in the rock mass, and might not be able to perceive the exposed rock in the last blasting round similarly \citep{Elmo2021,Erlend2023subjective} \label{item:perception}
    \item Safety and Accessibility Limitations: High-risk conditions and physical barriers often prevent thorough inspection of exposed rock, particularly in large double-track tunnels. Several of these tunnels are inspected from the floor at a safe location, leading to a poor assessment. \citep{Palmstrom2006,Elmo2021} \label{item:hazardous_inspection}
    \item Constraints on Quantification: Quantifying representative values for the newly exposed rock mass, such as the Rock Quality Designation (RQD), within the limited time frame of a fast-paced tunnel cycle presents significant challenges. The RQD metric, which involves measuring core pieces longer than 10 cm for every meter of rock, is tough to assess accurately under these conditions. \citep{Palmstrom2005,Pells2017rock, stille_classification_2003}. \label{item:quantification}
    \item Conservative Over-Supporting: Typically, the poorest rock mass conditions dictate the support for the entire blasted area, which can lead to unnecessary over-support in better-quality sections \citep{Palmstrom2006,Elmo2021} \label{item:not_finegrained} 
    \item Empirical Data Limitations: The original empirical data may not cover all geological conditions, construction geometries, or site-specific factors, resulting in an oversimplified approach to complex geologies \citep{Palmstrom2006,Elmo2021, Pells2017rock} \label{item:empirical}
    \item System Update Challenges: Updates to these systems (e.g. adding or adjusting features and site samples) are infrequent and labour-intensive, hindered by non-transparent design processes and inherent biases. The process involves trial and error to adapt input configurations to experienced stability at existing sites \citep{Elmo2021,Palmstrom2006,SEN2003269, stille_classification_2003}. \label{item:update} 
    \item Complex Rules for Exceptions: The systems incorporate complicated rules and factors that adjust classifications for different scenarios, which can lead to errors if not applied correctly, E.g. forgetting to multiply the Jn value (number of joint sets) in a junction with three, you might end up with a rock support class that is too low \citep{Palmstrom2006,Elmo2021, stille_classification_2003}. \label{item:complexity} 
    \item Visual Assessment Limitations: Current systems focus on visually assessable rock mass, neglecting the stability of rock outside the immediate tunnel profile, which is crucial for overall stability  \citep{Palmstrom2006,Elmo2021}. \label{item:visual_asessment}
    \item Non-existing advance rock mass assessment: The existing system cannot effectively assess the rock mass quality in front of the excavation, making it less useful for decision support on advance support and excavation method. You might say you can classify a drilled rock core from the face in RMR or Q-class, but such a process severely impacts the efficient tunnel factory and is only a point value. A system which describes advance support classes (face bolts, stability grouting, spiling bolts, etc.) from data ahead of the tunnel remains elusive \citep{Elmo2021,Palmstrom2006, stille_classification_2003}. \label{item:advance_support}
    \item Concervative Support: Rock support classes are defined by inspecting primarily stable conditions, majorly in civil infrastructure tunnels with a high safety factor and where the rock support is conservative, which may not reflect the actual stability needs \citep{Palmstrom2006,Elmo2021}. \label{item:concervative_support} 
    \item Mismatch in support classes: There are criticisms regarding how the combination of defined classes and linked rock support description can address the right type of rock support in general. \citep{Pells2007Limitations, Palmstrom2006,Ranasooriya2008}. \label{item:right_support}
    \item Failure Modes: Existing systems may not adequately consider various failure modes in their classifications, impacting the accuracy of rock support assessments \citep{Palmstrom2006,Elmo2021, stille_classification_2003} \label{item:failure_modes}
\end{enumerate}

Existing rock mass classification systems face several limitations: they are challenging to update, inherently conservative, subject to user bias, lack sufficient details, are unsuitable for forecasting, and do not assess the rock mass where support is most needed. Additionally, the conservative nature of the industry often places higher trust in the subjective assessments of experienced human experts over automated systems \citep{Elmo2021, Morgenroth2019}. This resistance to automation may hinder updates that could incorporate more systematic data collection. Transitioning rock mass assessments to a more transparent, reproducible, bias-free, and easily updatable \textit{decision support system} could address these issues effectively.

\subsection{The need for data-driven rock mass classification systems}

An accurate and objective understanding of rock mass stability is crucial to optimise rock support, blasting design, and excavation methods in tunnelling and mining. Rock materials are inherently heterogeneous, and the quality of the rock mass can vary significantly over short distances. However, historically, we have not been able to comprehensively describe this complexity to facilitate optimised decision-making in the fast-paced tunnel cycle. Consequently, rock mass quality is grouped into practical and meaningful target classes.

Existing classification systems are impractical in environments that are inaccessible to humans. These include hazardous areas of weak rock in current tunnelling projects, production mining environments operating close to a safety factor of 1.0, and extraterrestrial locations planned for future human bases. In these scenarios, automated systems are necessary for thorough rock mass assessments.

A finely tuned rock mass classification on a sufficiently small scale is essential for optimising decisions. Current broad-spectrum, largely subjective decisions are predominantly conservative, leading to the excessive use of steel and concrete for rock support. Moreover, there is a significant gap in our ability to assess rock mass quality ahead of the excavation, which is vital for planning advance support and selecting the appropriate excavation method. Addressing these challenges and the limitations outlined in Section~\ref{limitations} requires the development of new, more adaptive systems that can operate autonomously and provide accurate assessments in real time to enhance safety and efficiency in challenging and dynamic environments.

\subsection{Advances in data-driven classification outside the tunnel profile\label{sec:state-of-the-art}}

To date, no purely data-driven rock mass classification system for tunnelling and underground mining encompasses the following properties: (a) the capability to classify rock mass stability of the visually exposed rock mass and outside the tunnel profile, including the area ahead of the tunnel face; (b) independence from existing classification systems; (c) the use of comprehensive data, patterns and decisions from larger rock volumes than individual drillholes, such as entire blasting rounds or a 1 m slice of the tunnel; (d) practical classification of rock mass into stability classes that aid decision-making, beyond merely correlating mechanical properties like UCS, E-modulus, and single features.

The necessity for criterion (a) stems from assessing the rock mass's stability surrounding and ahead of the tunnel face. Criterion (b) avoids inheriting limitations from previous systems. Criterion (c) acknowledges that the rock mass quality can vary significantly over short distances. It is the combined signature of the rock mass for a larger volume that should be used to assess geotechnical risks and make practical decisions regarding rock support and excavation methods. Classifying from single drillholes might introduce so much noise and variations when making relevant decisions, based on the forecasting information from all the drillholes, that an extra interpretation step is needed to make a decision such as "Should I install spiling bolts or not?". Criterion (d) highlights that decisions on rock support and excavation methods cannot rely solely on single mechanical properties or features. 

While criteria (a), (c), and (d) are likely easier to accept and understand as necessary for new data-driven systems, criterion (b) requires further explanation. \citet{Yang2024} cautions against using rock mass classification labels from systems such as RMR, Q-system, and GSI for training ML models. These labels are subjective, derived from multiple input variables, and collected using varied practices worldwide. The empirical basis of these systems does not encompass the full range of rock mass behaviour, leading to models that may not generalise well across different geological contexts. Additionally, reliance on these classification systems could introduce biases and inaccuracies, as they are based on qualitative assessments lacking robust quantitative foundations \citep{yang_why_2022,hammah_does_2023,Yang2024}.

To our knowledge, the datasets derived from tunnel face seismics and Measure While Drilling (MWD) in hard rock tunnelling are the only ones providing the necessary spatial detail to effectively act as signatures of the rock mass beyond the tunnel profile. Several studies have linked high-resolution datasets to rock mass quality without explicitly aiming to develop a new data-driven classification system. We have categorised these studies into two groups based on their use of these datasets for forecasting purposes. Using geophysical data, \citet{Dickmann2021} and \citet{Dickmann2022} have automated the characterisation of rock mass into stability zones ahead of the tunnel and linked rock support to ground treatment using tunnel seismic prediction (TSP). \citet{Sapronova2021sparse} employed Principal Component Analysis (PCA) for dimension reduction and unsupervised clustering with the K-means algorithm to group similar data into clusters representing different geological conditions, subsequently labelled through supervised learning. These studies show promising results in characterising the rock mass ahead of the tunnel face. However, the development of a complete rock mass classification system is pending, and accuracy remains to be enhanced. Moreover, the use of geophysics at the excavation face involves human intervention. It could significantly affect the drill and blast cycle more than automatically collected MWD data.

\citet{Sapronova2024correlation} utilises correlation values between MWD features from single drillholes as feature vectors to predict Q-system classes, which contradicts points (b) and (c) regarding independence from existing systems and analysing larger rock volumes. Similarly, \citet{Hansen2024rockmassquality} employs statistically derived MWD values from all drillholes in a blasting round to predict Q-classes, yet still breaches point (b) by relying on existing classification systems. \citet{Fernandez2023} applies machine learning to single-hole MWD data to detect discontinuities using a calculated discontinuity index, thus contravening point (c) about using summarised data decisions from larger rock volumes. \citet{VanEldert2020} predicts Q-values and rock support from single-hole MWD data, violating points (b) and (c), although the study’s calculation and visualisation of a fracture index from single holes, segmented manually into different rock mass quality zones, moves towards a purely data-driven system. However, segmentation must be automated to reduce human bias, and explicit support classes must be linked to fracture indices. \citet{He2019} and \citet{Zhao2024} also use single-hole MWD data but use it to predict UCS with machine learning, infringing upon points (c) and (d), which call for broader data integration and practical decision-making utility. Lastly, \citet{Galende-Hernandez2018} clusters MWD data and links these to RMR values using expert-based fuzzy rules, breaching point (b) and violating the stated requirement to be purely data-driven by relying on human tuning in the approach.

Although conceptually promising, a common limitation in most existing studies is their base on small and site-specific datasets, which may lead to questionable generalisability\citep{Apoji2023}. \cite{Yang2024} describe how small and biased datasets severely affect claimed results in geotechnical ML modelling.

\subsection{Unsupervised machine learning for clustering and pattern detection in spatial data\label{sec:unsupervised_soa}}
In geoscience and related fields, unsupervised machine learning has been used to interpret, cluster, and identify patterns in spatial data and long-sequence data, as shown by recent studies. \citet{Sapronova2021sparse} demonstrated that cluster labels are often used in supervised learning models to predict classes for new data. Below, we present studies reflecting the current state of the art in this field. A common practice is the use of dimension-reduction techniques before clustering to enhance results.

\citet{Wiratama2022} applied unsupervised learning, specifically Gaussian Mixture Models (GMM) and K-Means, to classify facies in seismic images. GMM outperformed K-Means, particularly in identifying gas-rich sandstone formations, which were enhanced by PCA for dimensionality reduction. The study demonstrated that GMM provided more accurate and convincing facies classification results in the Carpathian Foredeep Basin.

\citet{Sherley2023} aimed to detect land changes from satellite images using unsupervised deep clustering techniques. The best results were achieved using deep embedded clustering (DEC) with Fuzzy-C-Means, yielding a silhouette score of 0.701. The methodology involved preprocessing satellite images, applying DEC and sparse autoencoder (SAE) models, and clustering with K-Means and Fuzzy-C-Means. The study concluded that deep clustering, particularly DEC with Fuzzy-C-Means, effectively addresses mixed pixel problems but requires high-quality data and computational resources.

\citet{Hasana2023A} investigated fire zone mapping in Indonesia using clustering algorithms and remote sensing. The best result was achieved with the ensemble UMAP (Uniform Manifold Approximation and Projection) + DBSCAN (Density-Based Spatial Clustering of Applications with Noise), yielding a silhouette score of 0.971 and a Davies-Bouldin index of 0.05. The study highlighted the method's effectiveness in identifying compact, well-defined fire-prone areas, contributing to improved fire hazard assessments.

\citet{Gare2023} developed a framework using UMAP-assisted HDBSCAN (Hierarchical DBSCAN) for analysing live cell calcium imaging data. The approach efficiently clustered Ca2+ spiking patterns with high accuracy, outperforming traditional methods. The study demonstrated significant correlations between cellular arrangement and Ca2+ oscillations, offering insights into biophysical mechanisms with a silhouette score indicative of robust clustering.

\citet{becht_dimensionality_2019} demonstrates the effectiveness of UMAP for dimensionality reduction in single-cell RNA sequencing data, a high-dimensional and complex dataset. The authors highlight how UMAP's ability to preserve both global and local data structures significantly improves the interpretation of cellular heterogeneity. Furthermore, the study shows that combining UMAP with HDBSCAN allows for identifying distinct cell populations which were not easily distinguishable with other methods. 

\citet{zhang_unsupervised_2024} generated a large-scale dataset of chromatin images from 560 tissue samples across 122 patients with various stages of breast cancer. The study utilized a Variational Autoencoder (VAE) to learn representations of cell states. 50 of the top VAE components were subsequently clustered using K-means clustering. Eight distinct morphological clusters were identified, which were consistent across all disease stages but varied in proportion. UMAP was used to visualize the latent space of the VAE, aiding in the interpretation of the clustering results and demonstrating how the clusters relate to different disease stages. The analysis revealed significant changes in tissue organization as the disease progressed, and the spatial organization of these cell states was predictive of disease stage and phenotypic category.

\subsection{Laying the groundwork for data-driven classification}

We can think of two natural strategies for optimising rock mass classification systems in a data-driven way: (a) employing modern data collection and learning techniques to improve the existing systems in various ways (e.g. improve assessment of input variables or extending the systems with new features), or (b) developing entirely new, purely data-driven systems. This study adopts the latter approach, hypothesising that rock mass can be intricately grouped, clustered, and classified using spatially extensive, high-resolution Measure While Drilling (MWD) data, which serves as a signature of the rock mass. Our objective is to address several limitations in existing systems in this context. MWD data is a cost-effective and easily retrievable data source in global tunnelling and mining operations \citep{VanEldert2017}, with the added advantage of not impacting the tunnel cycle. We analysed the natural clustering of MWD data, organised as tabular data samples for every meter of tunnel excavation across 15 hard rock tunnels, totalling 23,000 meters and involving approximately 500,000 blasting drillholes of infrastructure tunnel data. Such an extensive dataset reduces the risk of the results being overly sensitive to the dataset and improves generalisability compared to single-site datasets. The procedure was as follows: 

\begin{itemize}
    \item Information extraction involved calculating six statistical features from about 5000 values for each of eight MWD parameters, yielding 48 values in total. See Fig.\ref{fig:data_collection} for a visualisation of the process.
    \item To enhance clustering, we reduced the dimensionality of the 48-value vectors using nonlinear manifold learning techniques such as UMAP and linear Principal Component Analysis (PCA).
    \item We employed unsupervised machine learning techniques (HDBSCAN, Agglomerative Clustering, K-means) to identify natural groupings and structures within the data, creating clusters. We further optimised and explored the hyperparameters in these algorithms using multi-objective optimisation to ensure effective and meaningful clustering.
    \item We mapped and structured the clusters according to the physical properties of the rock mass, such as rock type and quality, and the distributions of key MWD parameters essential for rock mass assessment and clustering to determine whether the clustering meaningfully differentiated the rock mass.
\end{itemize}

The subsequent sections outline the dataset, methods used, results, and their analysis. The methodology section covers dimension reduction, clustering, decision metrics, organising experiments, hyperparameter optimisation, and linking clusters to physical properties of the rock mass. The discussion section explores the implications of these results. Finally, the conclusion and outlook sections summarise our findings and suggest directions for future research.

\section{Dataset and feature combinations\label{dataset}}

\begin{figure}
\centering
\includegraphics[width=1.0\textwidth]{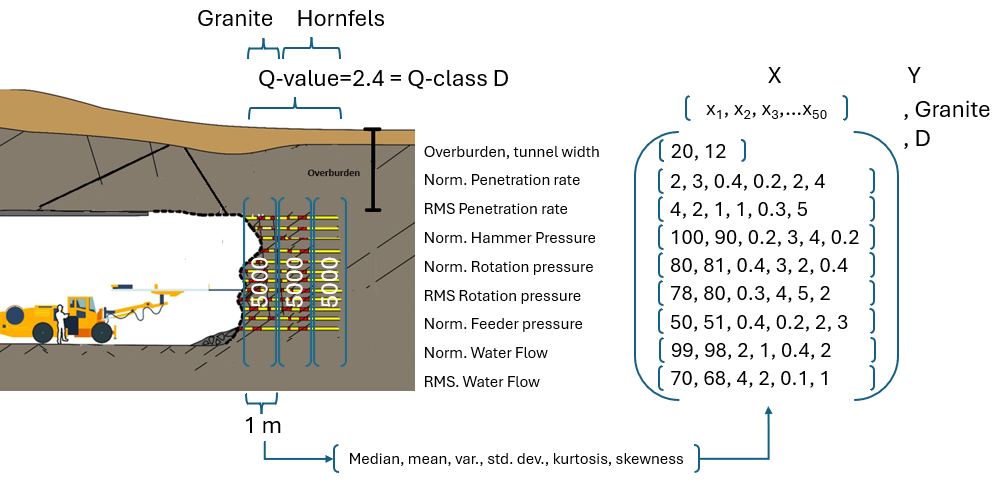}
\caption{Collection process for MWD-data and extraction of statistical information. X represents the features, and Y represents the labels. The two labels, Q-class D and rock type Granite are examples of values.}
\label{fig:data_collection}
\end{figure}

\begin{figure}
\centering
\includegraphics[width=0.8\textwidth]{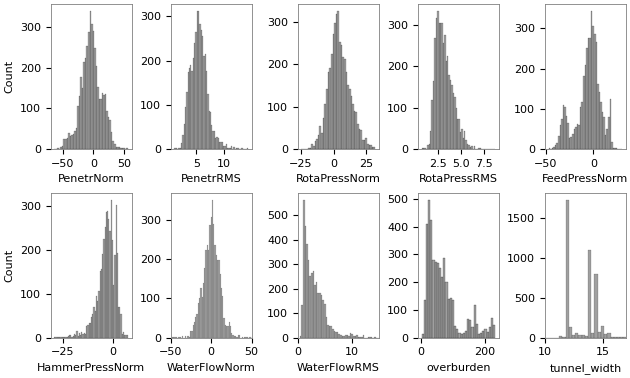}
\caption{Distribution of mean values for all MWD and geometric parameters in the dataset}
\label{fig:data_histograms}
\end{figure}

The dataset, detailed in \citet{Hansen2024dataset}, comprises 23,277 derived samples from approximately 500,000 drillholes in 15 hard rock tunnels. It features 48 MWD and two geometric parameters across 15 tunnels with varied geologies, originating from 4,202 blasting rounds. The primary rock types include Precambrian Gneisses, Permian Basalt and Granite, Permian Rhomb Porphyry, and Cambro-Silurian Shales, Limestone, and Claystone. Each sample also includes a label of the rock mass stability values Q-value and Q-class, computed using the Q-system\citep{Barton1974}. Five MWD parameters were logged and preprocessed using either normalisation, Root Mean Square (RMS) filtering, or both in separate operations, resulting in eight distinct preprocessed parameters. In abbreviated form, they are named PenetrNorm, PenetrRMS, RotaPressNorm, RotaPressRMS, FeedPressNorm, HammerPressNorm, WaterflowNorm, and WaterflowRMS. Table~\ref{tab:parameter_summary} lists the parameters and the abbreviations used in the study. For each 1 m tunnel section, which includes approximately 5,000 sensor values from 120 drillholes for a full face excavation, we computed six statistical metrics (mean, median, standard deviation, variance, skewness, and kurtosis) for each of the eight processed MWD parameters, resulting in 48 feature values. Combined with two geometric parameters (overburden and tunnel width), this yields a total of 50 feature values. Fig.~\ref{fig:data_collection} visualises the collection and statistical extraction process. Fig.~\ref{fig:data_histograms} displays the distribution of the median values for all MWD and geometric parameters. 

\begin{table}[h]
    \centering
    \caption{MWD-parameters with abbreviations and their normalised/filtered forms}
    \label{tab:parameter_summary}
    \begin{tabular}{lll}
    \toprule
    \textbf{Original parameter} & \textbf{Abbreviation for} & \textbf{Description} \\
    \textbf{name and unit} & \textbf{normalised/filtered form}& \\
    \midrule
    Penetration rate (m/min) & PenetrNorm & Normalised penetration \\
    Penetration rate (m/min) & PenetrRMS & RMS filtered penetration \\
    Rotation pressure (bar) & RotaPressNorm & Normalised rotation pressure \\
    Rotation pressure (bar) & RotaPressRMS & RMS-filtered rotation pressure\\
    Feeder pressure (bar) & FeedPressNorm & Normalised feeder pressure \\
    Hammer pressure (bar) & HammerPressNorm & Normalised hammer pressure \\
    Water flow (l/min) & WaterflowNorm & Normalised flush water flow\\
    Water flow (l/min) & WaterFlowRMS & RMS-filtered flush water flow\\
    \bottomrule
    \end{tabular}
\end{table}

Four distinct feature sets were utilised to analyse natural clustering tendencies in MWD data. Each set is identified by its name at the beginning of the sets described below.

\begin{itemize}
    \item \textbf{All}. All 48 MWD-parameters, plus the two geometric parameters overburden thickness and tunnel width (50 in total), are used to investigate in what way extra feature information, in addition to the MWD rock mass signature, impacts the clustering result.
    \item \textbf{MWD}. All 48 MWD parameters are used to inspect clustering results when only the MWD signature is used.
    \item \textbf{MWD\_rock}. Using domain and data knowledge to reduce the feature set to 30 by removing the 12 MWD-parameters dealing with WaterFlow, which might not impact the rock mass stability, and the correlated standard deviation parameter (correlated to variance).
    \item \textbf{MWD\_median}. The median of the 48 MWD parameters, giving eight features, is analysed to determine whether it adequately represents the rock mass.
\end{itemize}

\section{Methodology\label{methodology}}

The primary objective is categorising rock masses into groups with similar properties using high-resolution, spatially distributed MWD-data. These groups are then assessed for distinct physical property signatures, which make sense to use as a foundation for decision support systems in underground construction tasks such as rock support, excavation design or grouting effort. Furthermore, the clusters are investigated for alignment with the existing label sets of rock type and rock quality (Q-class). This study focuses on the initial critical step of exploring natural clustering in drilling data. 

We explored various feature sets, dimensionality reduction techniques, clustering algorithms, and hyperparameter tuning to determine effective clustering. Scaling the feature vector is crucial for both dimensionality reduction and clustering, ensuring uniform contribution of features to the analysis. We tested several scaling methods, including MinMaxScaler, StandardScaler, and RobustScaler \citep{scikit-learn}.

\subsection{Dimension reduction to improve clustering}

Dimension reduction before clustering improves computational efficiency and clarity by reducing noise and redundancy, thereby enabling clearer and more significant groupings in the reduced feature space \citep{bishop2006pattern, allaoui_considerably_2020}. To address the complexities of rock mass classification, we applied dimension reduction techniques to the predominately 48-50 value-long feature vectors, reducing them to 2 to 15 features. This is the second dimension reduction technique applied to the spatial data, as we previously reduced the thousands of MWD-values for 1 m tunnel sections by calculating statistical metrics (see Section~\ref{dataset} for details). Specifically, we utilised UMAP (Uniform Manifold Approximation and Projection) \citep{2018arXivUMAP} and PCA (Principal Component Analysis) \citep{HastieTibshiraniFriedman2009}. These methods aim to preserve and highlight local and global data structures, facilitating pattern recognition and grouping by clustering algorithms. Common in fields like bioinformatics, this sequential approach (e.g. UMAP + HDBSCAN) of reduction followed by clustering accommodates heterogeneous data typically seen in geotechnical engineering \citep{10.1371/journal.pcbi.1011288, allaoui_considerably_2020}. Aware of potential distortions in cluster relevance due to manifold learning \citep{Schubert2017Intrinsic}, we also perform clustering directly on the MWD-features and ensure a detailed inspection of their distributions in all experiments (not the dimension-reduced components) and alignment with existing dataset labels. Additionally, we employed dimension reduction with UMAP to visualise the high-dimensional data with clusters in 2D and 3D plots.

A successful hyperparameter configuration is significant for the algorithm performances. Table~\ref{tab:hyperparameter_dimred} lists and explains the key hyperparameters to tune for each dimension reduction algorithm, as recommended by \citet{scikit-learn, becht_dimensionality_2019, Jolliffe2016}.

\begin{table}[htbp]
\centering
\caption{Most important hyperparameters to tune for dimension reduction algorithms in the study.}
\label{tab:hyperparameter_dimred}
\begin{tabular}{p{1cm} p{2cm} p{8cm}}

\toprule
\textbf{Algo.} & \textbf{Param.} & \textbf{Description} \\
\midrule
PCA & \texttt{n\_components} & The number of principal components to keep. This parameter determines the dimensionality of the reduced feature space and is crucial for balancing the trade-off between dimensionality reduction and retaining variance. \\
     & \texttt{svd\_solver} & The algorithm used for computing the principal components. Choices include \texttt{auto}, \texttt{full}, \texttt{arpack}, and \texttt{randomized}. The selection can affect computational efficiency and accuracy, especially for large datasets. \\
\midrule
UMAP & \texttt{n\_neighbors} & Determines the size of the local neighbourhood UMAP considers when embedding. Smaller values capture local structure, while larger values capture global structure.\\
     & \texttt{min\_dist} & Controls the minimum distance between points in the low-dimensional space. Lower values preserve more of the local cluster density, while higher values lead to a more even spread of points. \\
     & \texttt{n\_components} & The dimensionality of the reduced space. Similar to PCA, this determines how many dimensions the data is reduced to, influencing the trade-off between reduction and information retention. \\
     & \texttt{metric} & The distance metric used to measure similarity between data points. Choices include \texttt{euclidean}, \texttt{manhattan}, \texttt{cosine}, and others. This parameter influences the shape of the embedded space and the preservation of data relationships. \\
\bottomrule
\end{tabular}
\end{table}

\begin{figure}
    \centering
    \includegraphics[width=1\linewidth]{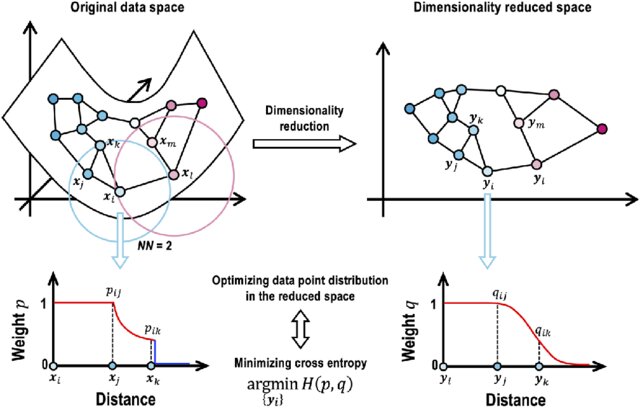}
    \caption{Schematic illustration of UMAP, adapted from \citet{oide_protein_2022}}
    \label{fig:umap_illustration}
\end{figure}

UMAP is a non-linear dimension reduction technique that preserves global and local structure of the data, making it particularly useful for clustering. Due to UMAP's significance for clustering and visualising non-linear data in biology and earth sciences \citep{allaoui_considerably_2020, Hasana2023A, Gare2023, oide_protein_2022}, its recent introduction in 2018\citep{2018arXivUMAP}, and its higher complexity compared to the mathematically simpler and more mature PCA technique, we have schematically described the method in Fig.~\ref{fig:umap_illustration}, adapted from \citet{oide_protein_2022}. The algorithm works by:

\begin{enumerate}
    \item Constructing a high-dimensional graph representation of the data (given as $N$ data points $x_i, i = 1, 2, \ldots, N$ in Fig. 3) based on a defined number ($NN$) of neighbouring points. For $NN = 2$, $x_i$ has the neighbouring points $x_j$ and $x_k$, measured by a distance metric (Euclidean and Manhattan are the most common).
    \item Optimising a low-dimensional version of this graph to be as structurally similar as possible. To preserve the relative data positions in the graph after the projection, we describe the weights between each point $x_i$ and its neighbours $x_j$ and $x_k$ using a function $p(x_i, x_j)$ or $p(x_i, x_k)$, depending on which pair of points we are considering. These weights are calculated using a probabilistic kernel function, where the weight $p_{ij}$ is highest (=1) when $x_j$ is the closest neighbour to $x_i$, and decreases as the distance between them increases, which is depicted by the curve in Fig.~\ref{fig:umap_illustration}. Similarly, $p_{ik}$ represents the weight between $x_i$ and its next nearest neighbour, $x_k$, and it also decreases as the distance increases, as shown in the same figure.
    
    In the figure, $p_{ij}$ is highest at point $x_j$, where $x_j$ is the closest neighbour of $x_i$, and decreases for further neighbours like $x_k$. To achieve the optimisation of a low-dimensional data representation that preserves as much of the original structure as possible, a cross-entropy function, $H$, defined in Eq.~\ref{eq:umap_crossentropy}, is minimised by adjusting the coordinates in the dimensionally reduced space $y_i, i = 1, 2, \ldots, m$.
\end{enumerate}


\begin{equation}
    H = \sum_{i,j} p_{ij} \log \frac{p_{ij}}{q_{ij}} + (1 - p_{ij}) \log \frac{(1 - p_{ij})}{(1 - q_{ij})}
    \label{eq:umap_crossentropy}
\end{equation}

The \textit{neighbouring points}, the \textit{distance metric between neighbours}, and the final \textit{number of chosen UMAP components} are among the most important hyperparameters to tune for your problem (see Table~\ref{appendix:hyperparameters} in Appendix~\ref{appendix:section_hyperparameters} for an overview). The default number of neighbouring points is 15. A higher number increases the preservation of the global structure but with a higher computational cost.

Conversely, PCA is selected for its efficiency and simplicity in linear dimension reduction. It works by identifying the directions (principal components) that maximise the variance in the data, which, for a dataset with a vector length of 50, is an effective strategy for reducing dimensionality while retaining as much information as possible. PCA's strength lies in its ability to provide a clear overview of the data's linear structure, making it a suitable linear counterpart to UMAP in our study. Using both UMAP and PCA, we can explore the data from different perspectives and potentially improve the performance of the clustering algorithms. 

While other dimension reduction techniques exist, both linear—such as Linear Discriminant Analysis (LDA) and Singular Value Decomposition (SVD)—and non-linear—such as t-SNE and autoencoders \citep{HastieTibshiraniFriedman2009,t-SNE,Goodfellow-et-al-2016}, we opted for UMAP and PCA due to their efficiency in handling the complexities of a big rock mass dataset, and their successful use in recent studies on spatial data, described in Section~\ref{sec:unsupervised_soa}. t-SNE, although powerful for manifold learning, is computationally intensive, challenging to tune and shows lower or similar performance to UMAP in several applications\citep{becht_dimensionality_2019}. LDA requires predefined class labels, which we aim to avoid as initial inputs. SVD is primarily utilised for matrix factorisation in contexts like natural language processing. Despite their capability for non-linear dimension reduction, Autoencoders demand intricate tuning and are complex to implement.

\subsection{Clustering approaches for rock mass data}

Following dimension reduction employing PCA, UMAP, and direct analysis without dimension reduction, we conducted clustering to understand the structural variations within the MWD signature of the rock mass. This approach is essential for identifying inherent groupings within the MWD data. We selected three distinct clustering algorithms based on their methodological diversity and applicability to our data type. Table~\ref{tab:hyperparameter_clustering} lists and explains the key hyperparameters to tune for each clustering algorithm, as recommended by \citet{scikit-learn, mcinnes2017hdbscan, hutter2019automated} and listed among the primary parameters in the algorithm documentation online. Proper hyperparameter configuration significantly impacts clustering outcomes. Some hyperparameters are mentioned in this section. More details of the tuning process are described in Section~\ref{hyperparameter_optimization}.

\begin{itemize}
    \item \textbf{K-means Clustering:} This algorithm was chosen for its efficiency in handling big data sets and simplicity, making it highly interpretable. K-means partitions the dataset into K distinct, non-overlapping clusters by minimising the variance within each cluster \citep{macqueen1967some}. This method is particularly effective for identifying spherical clusters in feature space. 
    \item \textbf{Agglomerative Clustering:} As a hierarchical clustering technique, this algorithm was employed to provide insights into the possible hierarchical structure of the dataset \citep{johnson1967hierarchical}. It progressively merges pairs of clusters that minimally increase a given linkage distance. Agglomerative clustering is useful for our study as it allows the examination of cluster structures at different scales. 
    \item \textbf{HDBSCAN:} Selected as an advanced density-based algorithm, HDBSCAN extends DBSCAN\citep{ester1996density} by converting it into a hierarchical clustering algorithm, introducing a stability-based cluster selection technique and simpler tuning. This choice is justified by its ability to handle variable cluster densities, which is crucial for datasets with complex spatial relationships like those found in geotechnical data \citep{campello2013density}. 
\end{itemize}

\begin{table}[htbp]
\centering
\caption{Most important hyperparameters to tune for clustering algorithms in the study.}
\label{tab:hyperparameter_clustering}
\begin{tabular}{p{1.5cm} p{2cm} p{8cm}}
\toprule
\textbf{Algo.} & \textbf{Param.} & \textbf{Description} \\
\midrule
HDBSCAN & \texttt{min\_cluster\_size} & The minimum size of clusters. This parameter determines the smallest group of points that should be considered a cluster, affecting the sensitivity to small clusters. \\
         & \texttt{min\_samples} & The minimum number of samples in a neighbourhood for a point to be considered a core point. Higher values make the algorithm more conservative in forming clusters.\\
         & \texttt{cluster} \texttt{\_selection} \texttt{\_epsilon}& The distance threshold for deciding whether points should be part of the same cluster. It helps in tuning the granularity of the clustering. \\
         & \texttt{metric} & The distance metric used to measure similarity between data points. Choices include \texttt{euclidean}, \texttt{manhattan}, \texttt{cosine}, and others. This parameter influences the shape and structure of the resulting clusters. \\
\midrule
K-means & \texttt{n\_clusters} & The number of clusters to form. This parameter is the most critical, as it directly controls the number of groupings in the data. \\
        & \texttt{init} & Method for initialisation of centroids. Options include \texttt{k-means++} and \texttt{random}. Proper initialisation can improve convergence speed and the quality of the final clusters.\\
        & \texttt{max\_iter} & Maximum number of iterations allowed for a single run. This parameter impacts the convergence behaviour and ensures the algorithm does not run indefinitely.\\
\midrule
Agglom.& \texttt{n\_clusters} & The number of clusters to find. This parameter sets the target number of final clusters after the hierarchical merging process. \\
Clustering   & \texttt{linkage} & The linkage criterion to use when merging clusters. Options include \texttt{ward}, \texttt{complete}, \texttt{average}, and \texttt{single}. The choice affects how the distances between clusters are calculated and, thus, the overall structure of the hierarchy. \\
             & metric& The metric used to compute the linkage. Common options are \texttt{euclidean}, \texttt{manhattan}, and \texttt{cosine}. It plays a crucial role in determining how clusters are formed based on the distances between data points. \\
\bottomrule
\end{tabular}
\end{table}

\begin{figure}[h]
    \centering
    \includegraphics[width=0.5\linewidth]{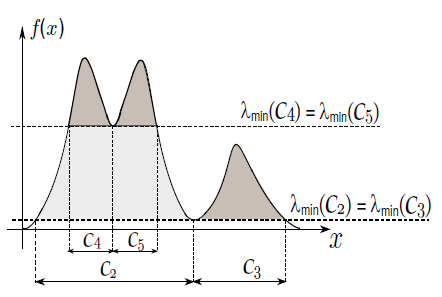}
    \caption{Illustrating density function and clusters for HDBSCAN, adapted from \citet{campello2013density}}
    \label{fig:hdbscan_density}
\end{figure}

\begin{figure}[h]
    \centering
    \includegraphics[width=0.5\linewidth]{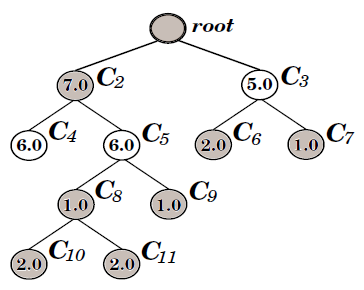}
    \caption{Illustrating the optimal selection of clusters for a cluster tree, adapted from \citet{campello2013density}}
    \label{fig:hdbscan_tree}
\end{figure}

As HDBSCAN is considered state-of-the-art in several studies, yet remains a newer and less familiar algorithm compared to the more mature and computationally simpler K-means and Agglomerative clustering, we provide additional explanations of HDBSCAN. Based on descriptions by \citet{campello2013density} and \citet{scikit-learn}, we explain HDBSCAN in an applied, less mathematically intensive manner. 

\textbf{Understanding Density}: HDBSCAN begins by estimating the density of data points within a dataset, guided by the \textit{distance metric} (e.g., Euclidean distance) used to calculate how close the points are to each other. The density is visualised as the curve \(f(x)\) in Fig.~\ref{fig:hdbscan_density}, where peaks represent areas of high density and potential clusters (e.g., \(C_2\), \(C_3\), \(C_4\), \(C_5\)). The algorithm identifies clusters by finding the minimum density level (\(\lambda_{\text{min}}\)) at which these clusters appear, influenced by parameters like \textit{minimum samples} (the minimum number of points in a neighbourhood to form a cluster) and \textit{minimum cluster size} (the smallest allowable cluster size).

\textbf{Building a Hierarchical Tree}: HDBSCAN constructs a hierarchical tree (shown in Fig.~\ref{fig:hdbscan_tree}) where each node represents a cluster formed at different density levels. This tree starts with the entire dataset as a single cluster at the root and splits into smaller clusters as density increases. The clusters \(C_2\), \(C_3\), \(C_4\), and \(C_5\) from Fig.~\ref{fig:hdbscan_density} are mapped onto this tree. The \textit{minimum cluster size} parameter plays a crucial role in determining when a split in the tree creates a valid cluster or should be treated as noise. Stability, reflected by the numbers next to the cluster labels, indicates how long a cluster persists as distinct.

\textbf{Selecting the Optimal Clusters}: From the hierarchical tree, HDBSCAN selects the most stable clusters using the \textit{cluster selection epsilon parameter}, which controls how sensitivity to density changes is handled when determining cluster boundaries. Stability is a measure of how long a cluster remains intact across different density levels (\(\lambda\)). In Fig.~\ref{fig:hdbscan_tree}, clusters like \(C_4\) and \(C_5\) are chosen because they exhibit higher stability, meaning they persist longer as the density threshold changes.

\textbf{Noise Identification}: Unlike K-means and agglomerative clustering, which assign every point to a cluster, HDBSCAN identifies samples that do not belong to any stable cluster as noise. This identification relies on their density relative to the \textit{minimum samples} and \textit{minimum cluster size} parameters. In Scikit-Learn's HDBSCAN implementation, these samples receive a label of -1. Points in low-density regions without significant clusters are excluded from the final clustering result and can often be interpreted as outliers. By adjusting the parameters—\textit{minimum samples}, \textit{minimum cluster size}, \textit{distance metric}, and\textit{ cluster selection epsilon}—HDBSCAN can effectively distinguishes between meaningful clusters and potential outliers in complex datasets. 

This ability is particularly beneficial in geotechnical data, where both noise points and small clusters can represent distinct geological phenomena. Noise samples, identified by HDBSCAN with a label of -1, typically represent isolated anomalous readings that do not fit well into any identified cluster. These points may indicate unusual or erratic conditions—for example, unstable rock masses or unexpected material changes—suggesting that these areas could require additional monitoring or investigation.

On the other hand, small clusters—clusters with a low number of samples—may represent niche geological phenomena that, while rare, share enough internal similarity to be grouped together. These small clusters could signify uncommon but stable features within the geological environment, such as a specific type of rock or a unique material property that is less frequent compared to the dominant geological formations.

Although noise samples and small clusters are not the same, both can provide critical insights: noise points often highlight potential anomalies or irregularities that may require attention, while small clusters might indicate specialised but consistent characteristics within the rock mass. By effectively isolating both types of data, HDBSCAN helps to focus attention on potential areas of concern or distinct features, thus enhancing the safety and reliability of geotechnical assessments.


The clustering approach does not rely on a predefined number of clusters; instead, it employs the Silhouette, Davies-Bouldin, and Calinski-Harabasz scores (see Section~\ref{evaluating_clustering} for a detailed description) to optimise the cluster count in a multi-objective optimisation described in Section~\ref{hyperparameter_optimization}. This method ensures that the optimal number of clusters is determined based on the data characteristics, avoiding bias from preset values. The implementation varies across algorithms; for instance, HDBSCAN does not require specifying a cluster number, whereas K-means and agglomerative clustering initially require a set number of clusters. However, as detailed later, the number of clusters is subsequently optimised, mitigating any potential issues from initial settings.

\subsection{Representation learning: a pipeline of unsupervised dimension reduction and clustering}
Representation learning refers to the process of automatically discovering and learning the representations or features from raw data that make it easier to perform tasks like classification, clustering, or regression\citep{Goodfellow-et-al-2016}. These learned representations can capture important patterns and structures in the data, often in a lower-dimensional space, making subsequent tasks more manageable. A study by \citet{zhang_unsupervised_2024} (see Section~\ref{sec:unsupervised_soa} for details) exemplifies successful representation learning by discovering breast cancer subtypes using a variational autoencoder (VAE) and subsequent clustering with the K-means algorithm. A VAE encodes input data into a probabilistic latent space, where each point represents a distribution, and decodes samples to generate similar data while preserving key features in a lower-dimensional space  \citep{Goodfellow-et-al-2016}. Below, we summarise our approach to representation learning using UMAP and HDBSCAN. 

Using UMAP, we transform high-dimensional sensor data (48-dimensional vectors) into a lower-dimensional space while preserving the essential structure and relationships within the data. This lower-dimensional representation captures important features or patterns relevant to the characteristics of the rock mass. After reducing the dimensionality, we apply HDBSCAN to identify clusters in the learned representations. These clusters represent different types of rock mass, characterised by similar patterns in the MWD sensor data.

The combination of UMAP and HDBSCAN enables the automatic discovery of meaningful representations from sensor data without the need for manually designed features. UMAP effectively learns a compact representation useful for identifying patterns, such as clusters of rock mass with similar properties. As we work with unlabeled data, this process constitutes \textit{unsupervised representation learning} aimed at uncovering hidden structures without relying on labelled examples.

\subsection{Evaluating clustering results\label{evaluating_clustering}}

The evaluation of clustering algorithms is a multifaceted process, presenting unique challenges compared to supervised learning. Unlike supervised learning, where precision and recall can directly measure performance, clustering evaluation must focus on the data's intrinsic structure without relying on predefined labels. This involves assessing whether the algorithm effectively identifies meaningful groups within the data that reflect some form of ground truth or underlying assumptions about data similarity \citep{scikit-learn}. The evaluation incorporates established clustering scores, qualitative assessments, and visual inspections, demonstrating the thoroughness and complexity of the process.

\subsubsection{Evaluating established clustering scores\label{sec:clustering_scores}}

Both label-based (external) and label-independent (internal) metrics were employed in the evaluation, focusing on the latter to maintain independence from existing labels (ref. the description of properties for data-driven rock mass classification systems in Section~\ref{sec:state-of-the-art}). \textit{External} metrics require true labels to assess the performance of the model. They are used to compare the clustering output to an externally provided true set of labels. External metrics evaluate how well the clustering has performed based on the known classification of the data. The external metrics included the Adjusted Rand Score and Adjusted Mutual Information Score. 

\textit{Internal} metrics do not require true labels and instead evaluate the quality of the clustering using the data itself. Internal metrics typically measure the compactness, separation, or density of the clusters formed by the model. The internal metrics used were the Silhouette coefficient (SC), Davis-Bouldin index (DBI), and Calinski-Harabasz index (CHI).

This balanced approach, incorporating both external and internal metrics, enables a thorough evaluation of the clustering algorithms. Below, we define the boundary values and briefly describe the interpretation of each score, starting with the external metrics.

Each feature vector in the dataset is assigned two labels: rock type and a rock mass stability class from the Q-system \citep{Barton1974}, as determined by face engineers during tunnel excavation. Label-based metrics such as the Adjusted Rand Index and Adjusted Mutual Information were utilised to evaluate the alignment between clustering outcomes and the labels for rock type and rock mass stability. These metrics provide insights into the relevance of clusters to actual rock mass properties, aiding the development of a data-driven rock mass stability classification system. Although the clustering aims to identify groups without strictly adhering to these labels, which were not used as targets in the optimisation process described in Section~\ref{hyperparameter_optimization}, assessing how well these labels align with the clusters is important. This alignment helps to validate the clusters against real rock mass properties and facilitates acceptance by the academic community.

\begin{itemize}
    \item \textbf{Adjusted Rand Score}:
    \begin{itemize}
        \item \textit{Boundary Values}: The score ranges from -1 to 1.
        \item \textit{Interpretation}: This metric assesses the similarity between two clusterings by considering all pairs of samples and counting pairs that are assigned in the same or different clusters in the predicted and true clusterings \citep{hubert1985comparing}. A score of 1 indicates perfect correspondence between the clustering labels and the true labels, implying an ideal match.
    \end{itemize}

    \item \textbf{Adjusted Mutual Information Score}:
    \begin{itemize}
        \item \textit{Boundary Values}: Ranges from 0 to 1.
        \item \textit{Interpretation}: Adjusted Mutual Information (AMI) is an adjustment of the Mutual Information (MI) score that accounts for chance. It measures the agreement of the two assignments, ignoring permutations \citep{vinh2010information}. A score of 1 denotes perfect agreement between the clustering labels and the true labels, adjusted for chance, which suggests a flawless clustering output.
    \end{itemize}
\end{itemize}

\noindent The internal metrics are described as follows:

\begin{itemize}
    \item \textbf{Silhouette Coefficient Score}:
    \begin{itemize}
        \item \textit{Boundary Values}: The Silhouette Score ranges from -1 to 1.
        \item \textit{Interpretation}: This metric evaluates the consistency within clusters by comparing the distance between objects within the same cluster to the distance to objects in the nearest cluster \citep{rousseeuw1987silhouettes}. A high value close to 1 indicates that objects are well-matched to their own cluster and distinct from neighbouring clusters, representing optimal clustering. In unsupervised learning, relying solely on one metric for evaluation is often insufficient. However, this study primarily utilises the Silhouette score to rank results, as it slightly more effectively differentiates between superior and inferior clustering outcomes, particularly when analysing cluster visualisations.
    \end{itemize}

    \item \textbf{Davies-Bouldin Index}:
    \begin{itemize}
        \item \textit{Boundary Values}: The score begins at 0 and has no predefined upper limit.
        \item \textit{Interpretation}: This index measures the average 'similarity' between clusters, where similarity is the ratio of within-cluster distances to between-cluster distances \citep{davies1979cluster}. Lower values indicate that clusters are well-separated and internally cohesive, with the optimal score being 0, suggesting minimal intra-cluster variation and maximal inter-cluster distinction.
    \end{itemize}

    \item \textbf{Calinski-Harabasz Index} (also known as the Variance Ratio Criterion):
    \begin{itemize}
        \item \textit{Boundary Values}: There is no upper limit, but higher scores indicate better clustering quality.
        \item \textit{Interpretation}: This score is calculated by measuring the ratio of the sum of between-clusters dispersion and within-cluster dispersion for all clusters \citep{calinski1974dendrite}. Essentially, a higher score signifies dense and well-separated clusters, which is considered indicative of a good clustering structure.
    \end{itemize}
\end{itemize}

Given the complexity of clustering evaluation, multi-objective optimisation and Pareto front analysis\citep{Deb2011multiobjective} are used for the internal metrics, to optimise and compare experimental results comprehensively. Section~\ref{hyperparameter_optimization} details the optimisation process, including a Pareto front analysis.

\subsubsection{Evaluation beyond standard clustering metrics}

Standard clustering metrics often overlook practical aspects of clustering quality, such as the distribution of cluster sizes and the presence of unclustered samples. To address these limitations, we included additional evaluations to ensure that the clustering outcomes align with both scientific expectations and practical needs:

\begin{itemize} 
    \item \textbf{The Gini index}, is a measure not typically associated with clustering but indicative of sample distribution uniformity within clusters \citep{HastieTibshiraniFriedman2009}. The Gini index, commonly applied in decision trees to evaluate node purity, varies from 0 to 1, where 1 signifies perfectly uniform cluster sizes. This aspect is not adequately captured by the metrics previously discussed. Despite favourable clustering scores, sample distribution within clusters can be highly uneven, which may not reflect the actual characteristics of the clusters. Experiments were excluded when sample distribution was excessively skewed, often with one cluster containing most samples while others had very few. A Gini index of 1 indicates equally sized clusters; however, values approaching 1 can also suggest suboptimal clustering performance, a potential issue in methods like K-means that prefer circular clusters \citep{scikit-learn}.
    \item \textbf{Number of unclustered samples.} The HDBSCAN algorithm detects outliers not included in clusters. Experiments were excluded if outliers represented more than 10\% of the samples, indicating suboptimal clustering. Non-clustered samples may result from incorrect configuration or genuine outliers in the dataset. Visualising the clusters can typically distinguish between these types.
    \item \textbf{Number of clusters and their sizes.} The practical value of clustering diminishes if the number of clusters is excessively high or too low (three or fewer). Experiments with such outcomes were discarded, particularly when a single cluster encompassed nearly all samples, a scenario not adequately addressed by existing cluster scores despite high scores.
    \item \textbf{The compactness of clusters}. The compactness of clusters indicates cluster quality and the potential for improvement by modifying configurations or adding features. For instance, a cluster that appears spread out and nearly divided into two suggests a need for refinement.
\end{itemize}

\subsubsection{Visual inspection of clustering}

Visual inspection of clustering, a crucial part of our evaluation process, was facilitated by Plotly Express \citep{plotly}. This tool allowed us to generate dynamic 2D and 3D plots, enabling detailed clustering examination through interactive manipulation such as rotation and zoom. Two plot variations were created: scatter points labelled by cluster value and rock type and rock mass stability class. Both plot types employ UMAP dimension reduction to 2 or 3 dimensions despite the typical use of more than ten components in the UMAP process for clustering. Additional details, such as rock type, are displayed when hovering over the scatter points. The 3D plots provide a unique perspective, allowing for the evaluation of clustering effectiveness and the relationship between clusters, which is hardly possible with other metrics. It is important to note that these visual assessments are limited to three dimensions. Thus, in cases where other metrics suggest superior clustering performance, discrepancies observed in the 3D plots do not necessarily indicate poor clustering if higher dimensions were visualised.

This comprehensive evaluation approach, combining established metrics, qualitative assessments, and visual inspections, facilitated a nuanced understanding of the clustering's effectiveness in classifying rock mass stability, laying the groundwork for a novel data-driven classification system.

\subsection{Organising the experimentation process\label{experimentation_process}}

The experimentation process in this study is structured as a pipeline comprising scaling, dimension reduction, clustering, and evaluation, implemented in Python and hosted on GitHub. Apart from UMAP \citep{2018arXivUMAP}, all components, including scalers, dimension reduction algorithms, clustering algorithms, and metrics, are sourced from Scikit-learn \citep{scikit-learn}.

The study conducted over a thousand experiments, primarily focusing on hyperparameter optimisation, with approximately 20\% undergoing detailed scrutiny. For efficient management and inspection of these experiments, we utilised mlflow \citep{mlflow2020} and hydra \citep{Yadan2019Hydra}. Mlflow was used to store and visualise the results of the experiments, along with the most important parameters and artefact files of features, true labels, and cluster values. Hydra was used to organise and store the detailed configurations of the experiments and to make it easy to run the experiments with different configurations. All configuration values were quality-checked using Pydantic \citep{Pydantic} in each experiment. The experiments were run in a makefile, which is a simple way to run the experiments in an efficient and reproducible way \citep{GnuMake}. Result plots and compilations were generated reproducibly by retrieving data from mlflow using unique experiment IDs. For further insights into scientific ML-based reproducible and transparent experimentation, see \citep{Hansen2024rocktype}.

\subsection{Hyperparameter optimisation \label{hyperparameter_optimization}}

Optimising hyperparameters is essential for effective clustering, given the sensitivity of results to these settings in dimension reduction and clustering algorithms. Proper hyperparameter configuration significantly impacts clustering outcomes. As outlined in Section~\ref{evaluating_clustering}, evaluating multiple metrics is crucial in unsupervised clustering. \citet{Hasana2023A} further emphasises the importance of handling irregular cluster shapes and optimising multiple metrics simultaneously to achieve accurate clustering results in their pipeline optimisation using UMAP and DBSCAN. In this study, the Silhouette score, Davies Bouldin index, and Calinski Harabasz index were selected as objective scores for optimisation. These internal clustering metrics (ref. Section~\ref{sec:clustering_scores}) keeps the optimisation process independent from the existing rock mass classification systems (ref. Section~\ref{sec:state-of-the-art}). We employed Bayesian multi-objective optimisation, using the Optuna package \citep{akiba2019optuna}, to adjust the hyperparameters in several stages. Bayesian optimisation is a probabilistic model-based optimisation method that is particularly useful for optimising expensive-to-evaluate functions.   

For hyperparameter sampling, we adopted the multi-objective version of the Tree-structured Parzen Estimator (MOTPE) \citep{Ozaki2022MOTP}, which is a specific implementation of the broader concept of Bayesian multi-objective optimisation. MOTPE effectively approximates a Pareto front and is more efficient than evolutionary samplers like NSGA-II, which directly optimise a Pareto front. In the context of multi-objective optimisation, a Pareto front represents the set of solutions considered non-dominated, meaning no solution can improve one objective without worsening at least one other objective \citep{Deb2011multiobjective}. Each point on the Pareto front corresponds to a solution where all objectives are balanced optimally. This concept is fundamental in multi-objective optimisation as it visually and numerically illustrates the trade-offs between competing objectives, helping decision-makers choose the most suitable solutions according to their preferences. Due to its significance in this study, we provide details on the optimisation process.

\textbf{Approximating the Pareto Front:} MOTPE extends the traditional TPE for multiobjective optimisation by defining vector dominance and using a nondomination ranking system. Observations are split into two groups via density functions based on their dominance relations to a reference set \( Y^* \), ensuring \( p(y \succ Y^* \cup y \parallel Y^*) = \gamma \). Additionally, the Hypervolume Subset Selection Problem (HSSP) is utilised to select observations that maximise the hypervolume indicator, facilitating an effective and diverse approximation of the Pareto front.

\textbf{Selection of Pareto Optimized Experiments:} In MOTPE, the selection of experiments is guided by the Expected Hypervolume Improvement (EHVI) criterion, which prioritises candidate solutions based on their potential to extend the current Pareto front. During each optimisation iteration, candidates are sampled from a distribution, and the one with the highest EHVI is evaluated. This greedy approach ensures continuous refinement of the Pareto front by balancing exploration and exploitation throughout optimisation. MOTPE builds models to estimate the probability distribution of objective values given the hyperparameters and then samples new hyperparameters from areas expected to yield better objective values. The most important parameters were defined for each algorithm (HDBSCAN, K-means, UMAP, etc.), and ranges of values were set up for the process of choosing new parameters. See Table~\ref{tab:hyperparameter_dimred} and \ref{tab:hyperparameter_clustering} for a description of these parameters. Each iteration follows these steps: 

\begin{enumerate}
\item \textbf{Sampling.} The MOTP sampler samples a new set of parameters from the hyperparameter space. This sampling is influenced by the past performance of parameter sets, aiming to explore regions with potentially better trade-offs between objectives.
\item \textbf{Evaluation.} Using a parametric pipeline including scaling, dimension reduction and clustering, a new set of hyperparameters was tested in each iteration, and the three objective metrics were reported from each run. According to each metric' optimised goal, the target metrics were set to be (ref. order above) maximised, minimised and maximised.
\item \textbf{Updating.} Based on the outcomes of the evaluations, Optuna updates its understanding of the hyperparameter space. For TPE, it involves updating the internal probabilistic models.
\item \textbf{Selection:} In the next iteration, Optuna uses the updated information to sample new parameters again, effectively iterating towards better solutions over time. In multi-objective optimisation, the 'better' solution must consider the trade-offs between competing objectives.
\end{enumerate}

In choosing the final best set of hyperparameters for each pipeline, this study focused on the Silhouette score and the Calinski Harabasz index, when several Pareto optimal solutions were found.

\subsection{Characterising the clusters\label{characterising_clusters}}

This study primarily investigates the natural clustering of MWD-data to establish a foundation for a data-driven rock mass classification system. The second step involves examining the physical properties of the rock mass within each cluster. Although this research initiates the process and demonstrates promising results and guidelines, further investigation into the properties of each cluster and refinement of cluster compositions are necessary to enhance the system's industrial applicability. That is a task for future studies. The properties of each cluster have been analysed in multiple ways across three high-performing experiments:

\begin{itemize}
    \item The cumulative distribution function (CDF) plots for three important MWD-parameters—Normalised penetration, feeder pressure, and rotation pressure (torque)—are presented for each cluster, as identified in studies by \citet{Schunnesson1998} and \citet{Navarro2018}. These parameters, which intuitively reflect the physical properties of the rock mass (e.g., higher penetration suggests softer rock, and higher feeder pressure indicates better rock quality), are analysed to determine if clustering effectively discriminates between rock masses with distinct physical characteristics. A good spread of distinct distribution suggests effective separation, whereas overlapping distributions imply less distinction.
    \item A table is provided that lists the median values and sample counts of the three MWD-parameters across clusters, sorted by increasing normalised penetration.
    \item To investigate alignment with existing classification systems, rock types (a classification of a certain aggregate of mineral grains) and Q-class labels are assigned to each cluster based on a majority vote (the most frequent label for samples in a cluster). If clusters predominantly share the same rock type or Q-class label, it indicates poor alignment. Alignment is also demonstrated visually by labelling samples with these classes in 3D cluster plots. We emphasise that \textit{this study does not aim to align the new classification approach with existing systems}, thereby breaching criterion (b) in Section~\ref{sec:state-of-the-art}. However, the demonstrated alignment provides insights into cluster properties and the existing systems' alignment with this new data-driven approach. Some alignment with existing systems might enhance the acceptance of this new approach by the industry and academic community and facilitate mapping actions, such as rock support measures, to each class. Notably, the rock type label is less subjective and easier to label correctly on-site than the Q-class.

\end{itemize}

\section{Results and analysis\label{results}}

This section is structured into three parts. First, it presents the outcomes of the hyperparameter optimisation process. Next, it details the clustering outcomes for different combinations of feature sets, dimension reduction techniques, and clustering algorithms and analyses the emerging patterns. The analysis is grouped into configurations leading to good or bad/questionable clustering. Finally, it explores the physical properties of each cluster and their association with tunnel decisions.

\subsection{Optimising the pipeline of dimension reduction and clustering}\label{optimizing_pipeline}

\begin{figure}[h]
    \centering
    \includegraphics[width=1.0\textwidth]{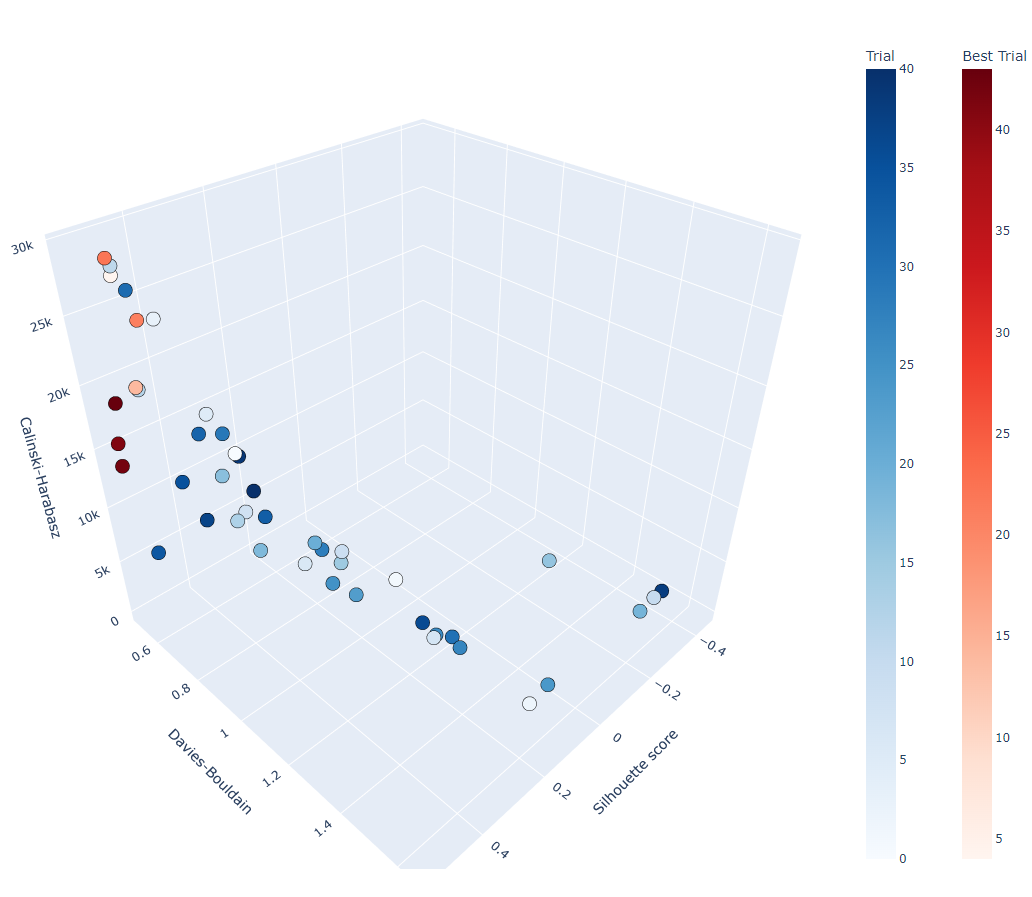}
    \caption{Pareto plot of multi-objective optimization for dimension reduction with UMAP and clustering with the agglomerative clustering algorithm. An interactive version of the plot is available in the digital version of the paper.}
    \label{fig:pareto_optimization_umap_agglomerative}
    \end{figure}

Approximately 50 experiments were required for three cluster metrics to converge, enabling the identification of optimal hyperparameters for each pipeline. Optimisation proved essential for generating informative clusters as default algorithm settings were inadequate. The goal was to maximise the Silhouette and Calinski-Harabasz scores while minimising the Davis-Bouldain index. The best hyperparameters identified in each run were logged in mlflow and applied in the final clustering experiments. The results of this hyperparameter optimisation are displayed in an interactive Pareto plot using a Plotly function \citep{plotly}, implemented in the Optuna package \citep{akiba2019optuna}. The plot in Fig.~\ref{fig:pareto_optimization_umap_agglomerative} shows the trials for an 'MWD' featureset optimisation involving UMAP and Agglomerative Clustering. Increasing trial numbers are indicated by progressively darker shades of blue, and improved Pareto optimal solutions are marked in darker reds. This presentation method effectively compares experiment outcomes against all objectives and illustrates the typical scatter development from the optimisation process. The MOTPE-sampler methodically explores parameter configurations, gradually converging on a Pareto front of optimal solutions.

The optimisation process, along with reporting and plotting, was executed for each distinct combination of feature set, dimension reduction technique, and clustering algorithm, leading to a comprehensive database of experiments tracked in mlflow. The four most effective experiments for a combination, as determined by the Pareto optimisation process, had their hyperparameters and unique IDs recorded. Each experiment was then thoroughly examined for its clustering results, and at least one experiment per configuration is detailed in two consolidated tables, Table~\ref{table:clustering_results} and Table~\ref{table:clustering_metrics} , in Section~\ref{comparing_clustering}. The optimised hyperparameters for three of these experiments are listed in Table~\ref{appendix:hyperparameters} in Appendix~\ref{appendix:section_hyperparameters}. Any parameters not specified revert to the algorithm's default settings.

Optimisation of hyperparameters significantly enhanced the clustering results, as evidenced by improved scores in the clustering metrics for experiments 0, 3, and 7, detailed in Table~\ref{table:clustering_results}. The SC, DBI, and CHI metrics showed marked improvements, particularly in pipelines using the HDBSCAN algorithm. Through optimisation, HDBSCAN pipelines transitioned from 25\% unclustered samples and over 800 clusters to effective clustering. In contrast, pipelines with the agglomerative clustering algorithm exhibited scores for default parameters that were closer to those of optimised parameters. Additionally, the occasional higher CHI scores for default configurations underscore the necessity of optimising multiple metrics and comprehensively evaluating various approaches.

\subsection{Comparing clustering results for different experiments}\label{comparing_clustering}

In Table~\ref{table:clustering_results} and~\ref{table:clustering_metrics}, the results of the best-performing experiments, trained with optimised parameters, are presented. The results are grouped by featureset and then sorted after the Silhouette score.

\begin{table}
    \centering
    \caption{Summary of clustering results for four different feature sets, grouped by feature sets. Scores for default algorithm parameters in parenthesis}
    \label{table:clustering_results}
    \begin{tabular}{cccccccccc}
    \toprule
    Id & Feature & Num. & Dim. & Cluster & Num. & Num. dim. & Num. not & Gini \\
    & set & features & red. & alg. & clusters & red. & clustered & index \\
    & & & alg. & & & comp. & samples & \\
    \midrule
    0 & all & 50 & umap & hdbscan & 9(956) & 12(2) & 0(6140) & 0.5(0.6) \\
    1 & all & 50 & umap & hdbscan & 9 & 15 & 23 & 0.5 \\
    2 & mwd & 48 & umap & aggl.\ clust. & 6 & 7 & 0 & 0.55 \\
    3 & mwd & 48 & umap & aggl.\ clust. & 7(6) & 6(2) & 0(0) & 0.57(0.24) \\
    4 & mwd & 48 & umap & hdbscan & 5 & 12 & 23 & 0.66 \\
    5 & mwd & 48 & umap & kmeans & 7 & 10 & 0 & 0.2 \\
    6 & mwd & 48 & umap & kmeans & 3 & 4 & 0 & 0.36 \\
    7 & mwd & 48 & umap & hdbscan & 13(838) & 3(2) & 38(6053) & 0.63(0.63) \\
    8 & mwd & 48 & umap & hdbscan & 13 & 12 & 1195 & 0.44 \\
    9 & mwd & 48 & pca & kmeans & 10 & 2 & 0 & 0.22 \\
    10 & mwd & 48 & umap & hdbscan & 6 & 2 & 22 & 0.69 \\
    11 & mwd & 48 & pca & hdbscan & 3 & 5 & 1842 & 0.59 \\
    12 & mwd & 48 & None & hdbscan & 3 & --- & 1 & 0.67 \\
    13 & mwd\_rock & 30 & pca & kmeans & 6 & 2 & 0 & 0.22 \\
    14 & mwd\_rock & 30 & umap & hdbscan & 11 & 4 & 175 & 0.68 \\
    15 & mwd\_rock & 30 & umap & hdbscan & 9 & 15 & 1014 & 0.64 \\
    16 & mwd\_median & 8 & umap & hdbscan & 6 & 11 & 23 & 0.74 \\
    \bottomrule
    \end{tabular}
\end{table}

\begin{table}
    \centering
    \caption{Summary of clustering metrics grouped by feature set and ordered by Silhouette score. Adjusted Rand score is calculated for the label rock quality for featureset `all', label rock type for other featuresets. Scores for default algorithm parameters in parenthesis}
    \label{table:clustering_metrics}
    \begin{tabular}{cccccccc}
    \toprule
    Id & Feature & Dim. & Cluster alg. & Silhouette & Davies & Calinski & Adjusted \\
    & set & red. & & & Bouldin & Harabasz & rand \\
    \midrule
    0 & all & umap & hdbscan & 0.65(0.15) & 0.39(1.61) & 68134(64776) & 0.02(0.002) \\
    1 & all & umap & hdbscan & 0.65 & 0.39 & 61432 & 0.02 \\
    2 & mwd & umap & aggl.\ clust. & 0.54 & 0.54 & 14031 & 0.3 \\
    3 & mwd & umap & aggl.\ clust. & 0.52(0.48) & 0.5(0.72) & 18316(31696) & 0.33(0.27) \\
    4 & mwd & umap & hdbscan & 0.45 & 0.44 & 9343 & 0.08 \\
    5 & mwd & umap & kmeans & 0.45 & 0.91 & 22210 & 0.25 \\
    6 & mwd & umap & kmeans & 0.45 & 0.82 & 15279 & 0.17 \\
    7 & mwd & umap & hdbscan & 0.41(0.11) & 1.17(1.46) & 13265(94512) & 0.47(0.04) \\
    8 & mwd & umap & hdbscan & 0.37 & 1.12 & 20866 & 0.24 \\
    9 & mwd & pca & kmeans & 0.32 & 0.88 & 13380 & 0.09 \\
    10 & mwd & umap & hdbscan & 0.3 & 0.46 & 7627 & 0.11 \\
    11 & mwd & pca & hdbscan & 0.23 & 2.29 & 974 & 0.01 \\
    12 & mwd & None & hdbscan & 0.74 & 0.35 & 307 & 0.0 \\
    13 & mwd\_rock & pca & kmeans & 0.37 & 0.85 & 17916 & 0.04 \\
    14 & mwd\_rock & umap & hdbscan & 0.26 & 1.33 & 7635 & 0.29 \\
    15 & mwd\_rock & umap & hdbscan & 0.18 & 2.76 & 9536 & 0.25 \\
    16 & mwd\_median & umap & hdbscan & 0.36 & 0.42 & 6142 & 0.17 \\
    \bottomrule
    \end{tabular}
\end{table}

\subsubsection*{Configurations leading to good clustering.}
We found that using a min-max scaler, scaling to a 0-1 range worked better than other scalers.

Considering the internal metrics (Silhouette score, Calinski Harabasz and Davies Bouldain index) and visual appearance in plots, the pipeline with dimension reduction with UMAP and clustering with HDBSCAN or Agglomerative clustering works equally well, but only for the feature sets `All' (50 features) and `MWD' (48 features). However, in comparing HDBSCAN (experiment 7) and Agglomerative clustering (experiment 3), we want to point out the higher Adjusted Rand Score for HDBSCAN in experiment 7. Visualising the clusters in 3D plots in Fig.~\ref{fig:experiment_0_all}a (more detailed in Fig.~\ref{fig:appendix_all_3D} in Appendix~\ref{appendix:section_cluster_performance}) for experiment 0, using the `All' (MWD and geometric features) featureset, Fig.~\ref{fig:experiment7_mwd}a for experiment 7 (more detailed in Fig~\ref{fig:appendix_mwd_experiment_7_clustercolour} and \ref{fig:appendix_mwd_experiment_7_detailed_geology} in Appendix~\ref{appendix:section_cluster_performance}) and experiment 3 in Fig.~\ref{fig:mwd_3D_agglomerative} in Appendix~\ref{appendix:section_cluster_performance} for the `MWD' featureset, demonstrates a near perfect match with the plotted clusters. Notably, a dimension reduction of each feature set with UMAP to three components has been used to facilitate plotting the scatter points, accounting for the differences in cluster scatter points in Fig.~\ref{fig:experiment7_mwd} and Fig.~\ref{fig:mwd_3D_agglomerative} in Appendix~\ref{appendix:section_cluster_performance} to be different from the ones in Fig.~\ref{fig:experiment_0_all}. In Fig.~\ref{fig:experiment_0_all} and \ref{fig:experiment7_mwd}, we have plotted the clustering results with three types of labelling, the cluster class in Subfigure \textbf{a}, the Q-class in Subfigure \textbf{b}, and rock type in Subfigure \textbf{c}. We want to point out that including the rock type label in Fig.~\ref{fig:experiment_0_all} is somewhat misleading since experiment 0 includes features which should not be related to rock type --- the geometric features. However, we have included the subfigure for comparison to the same kind of subfigure in Fig.~\ref{fig:experiment7_mwd}, discussed in a later section.

In experiment 10, we employed two UMAP components for dimension reduction and visualisation. The resulting plot, shown in Fig.~\ref{fig:2Dscatter_cluster_2Dumap_hdbscan_mwd}, displays a clear alignment between clustering results and visible clusters. However, the pattern of one large cluster alongside several smaller ones may not accurately represent the characteristics of this rock mass data. This assertion is further supported by lower performance scores.

\begin{figure}
    \centering
    \begin{minipage}[b]{\textwidth}
        \centering
        \includegraphics[width=0.5\textwidth,height=5cm]{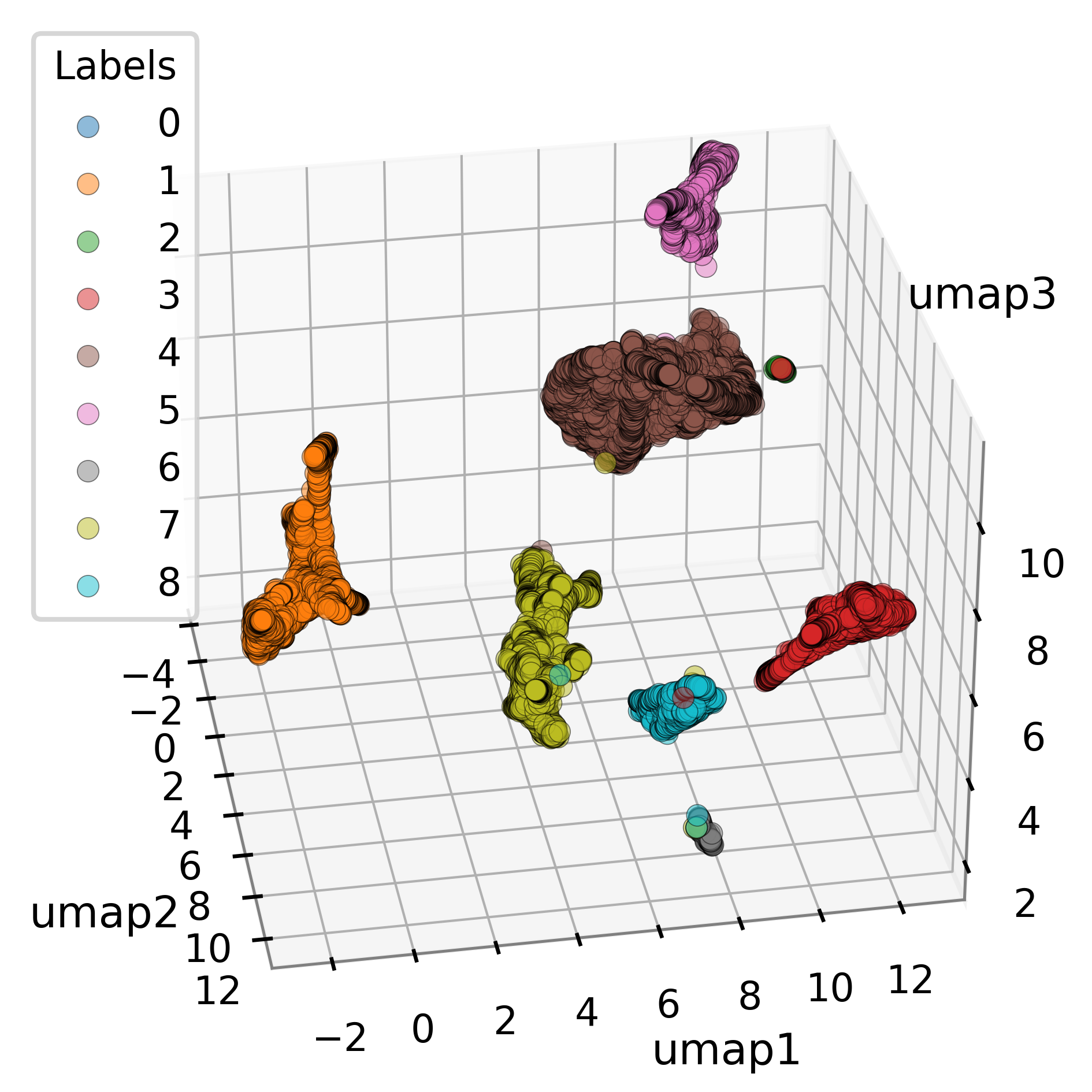}
    \end{minipage}
    
    \vspace{0.3cm} 

    \begin{minipage}[b]{\textwidth}
        \centering
        \includegraphics[width=0.5\textwidth,height=5cm]{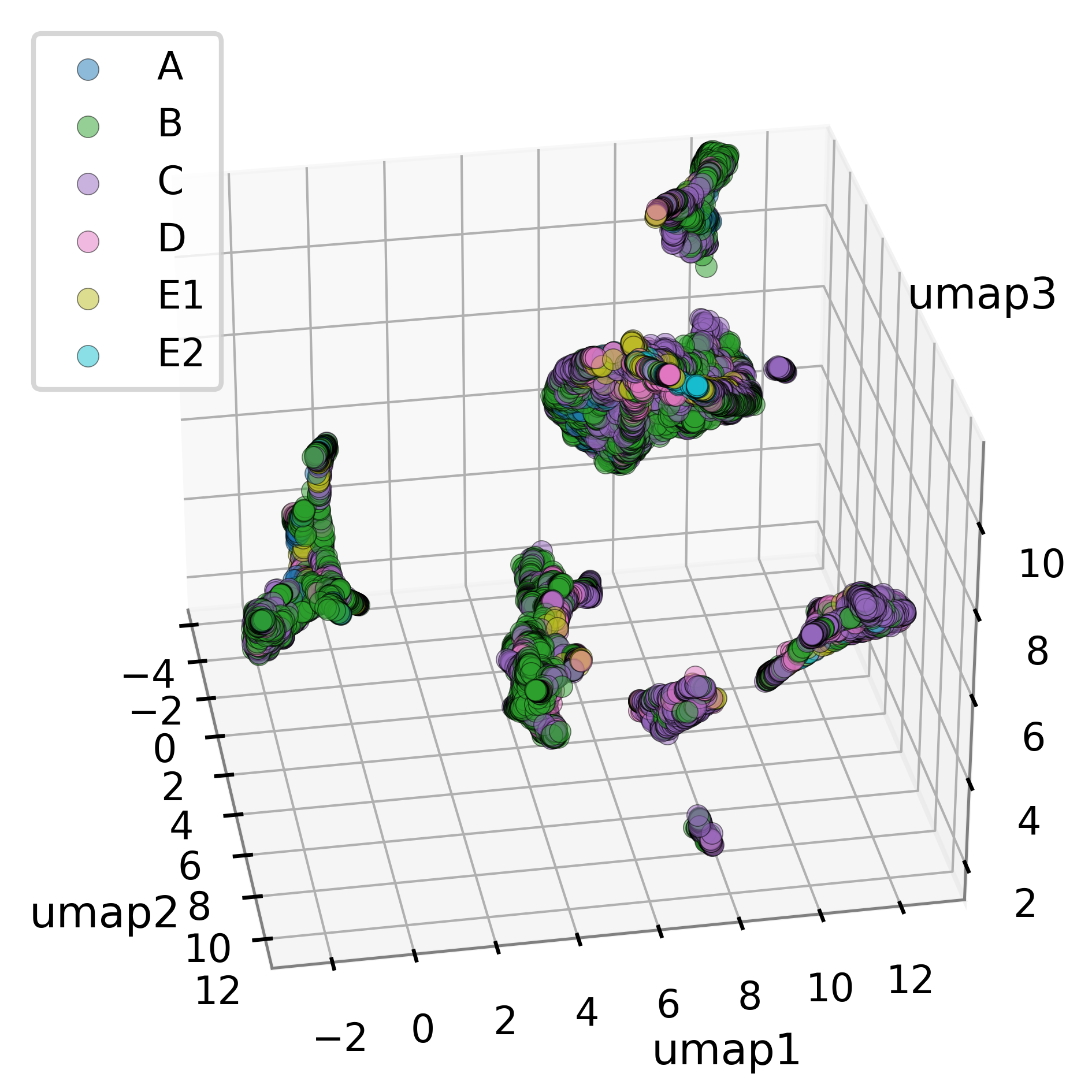}
    \end{minipage}
    
    \vspace{0.3cm} 

    \begin{minipage}[b]{\textwidth}
        \centering
        \includegraphics[width=0.5\textwidth,height=5cm]{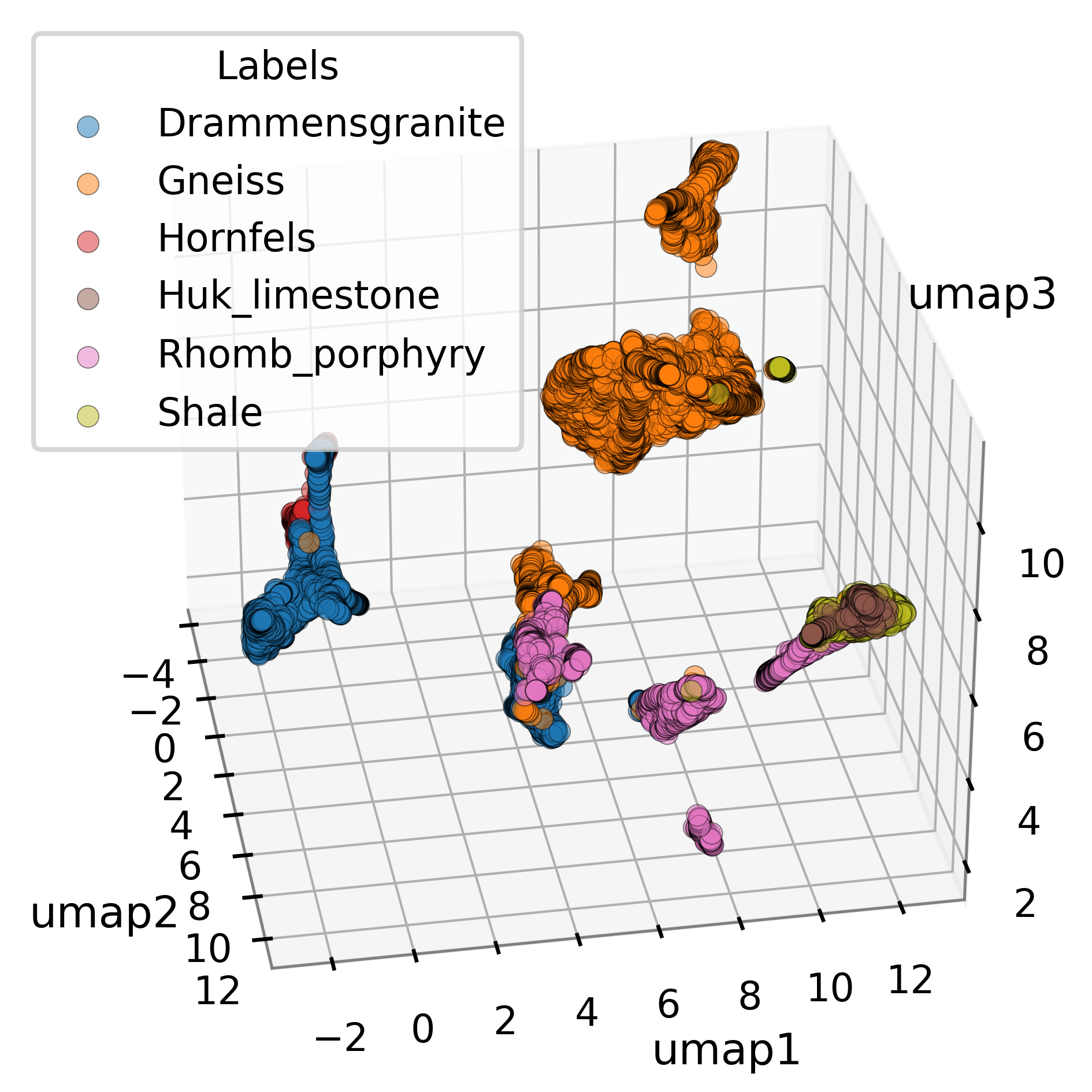}
    \end{minipage}
    
    \caption{Experiment 0. Clustering of MWD data using UMAP for dimension reduction and HDBSCAN for clustering, \textbf{a}  Clusters coloured by cluster number, \textbf{b} Clusters coloured by rock quality (Q-class), \textbf{c} Clusters coloured by rock type. An interactive version of the plot is available in the digital version of the paper.}
    \label{fig:experiment_0_all}
\end{figure}

\begin{figure}
    \centering

    \begin{minipage}[b]{\textwidth}
        \centering
        \includegraphics[width=0.5\textwidth,height=5cm]{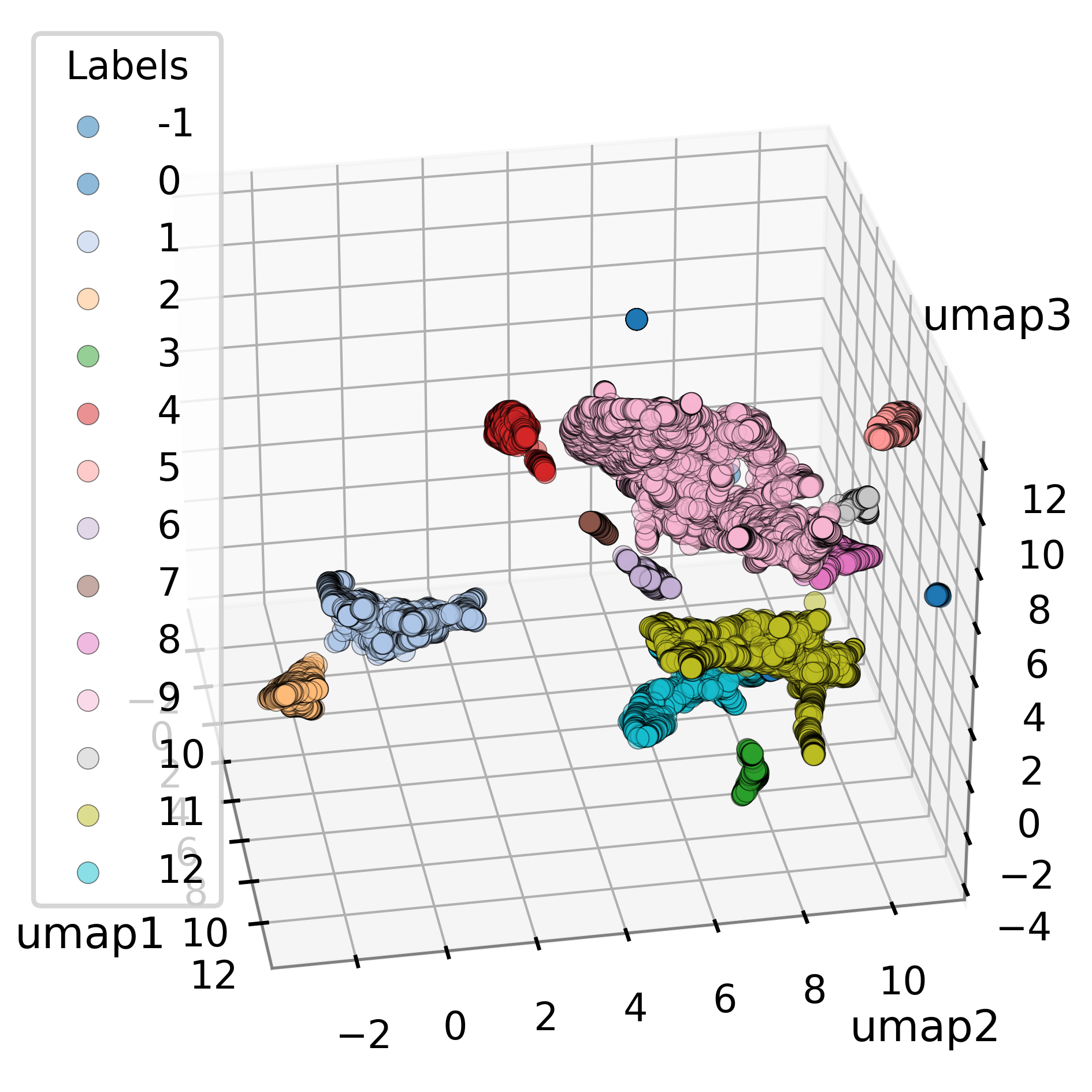} 
    \end{minipage}
    
    \vspace{0.3cm}
    
    \begin{minipage}[b]{\textwidth} 
        \centering
        \includegraphics[width=0.5\textwidth,height=5cm]{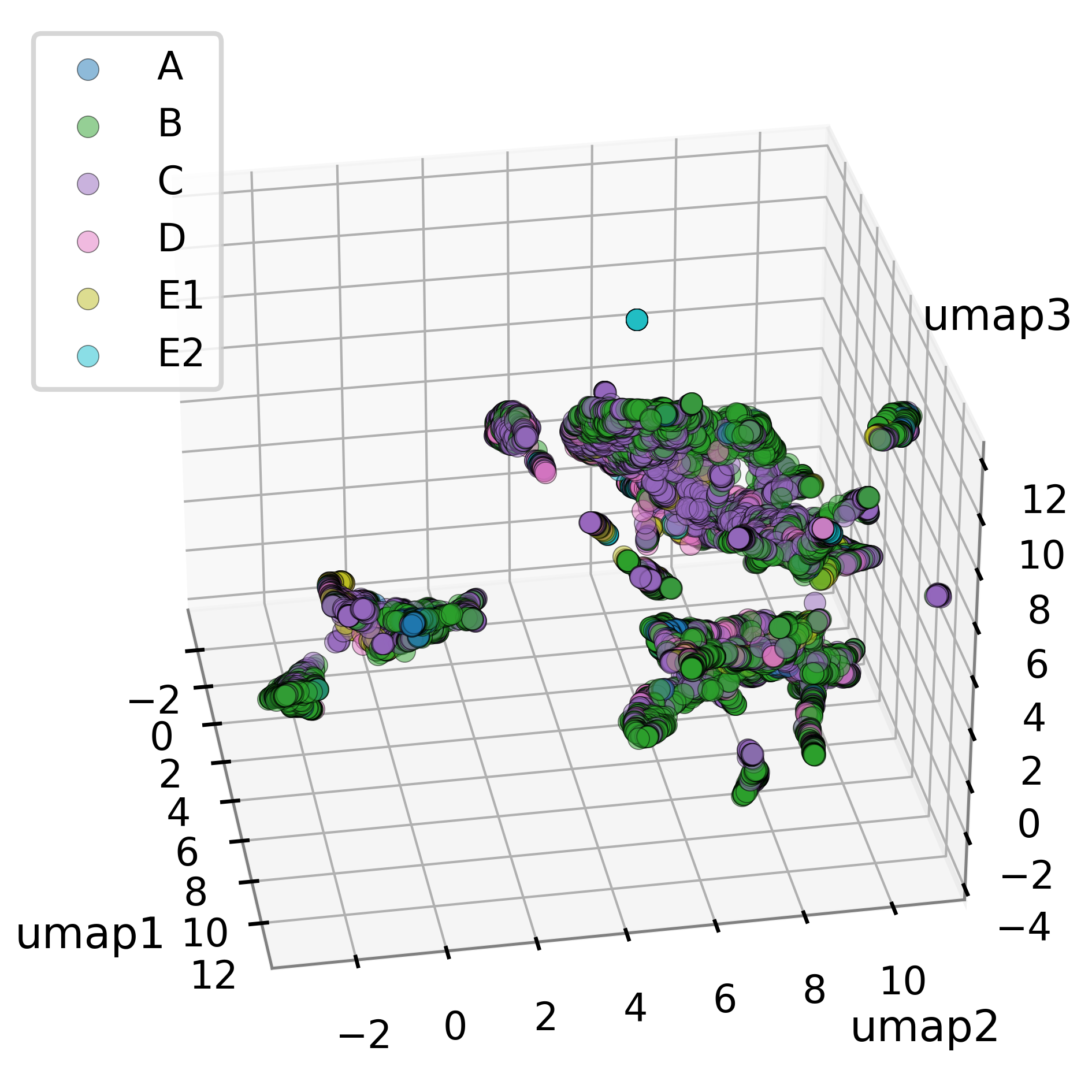} 
    \end{minipage}
    
    \vspace{0.3cm}

    \begin{minipage}[b]{\textwidth} 
        \centering
        \includegraphics[width=0.5\textwidth,height=5cm]{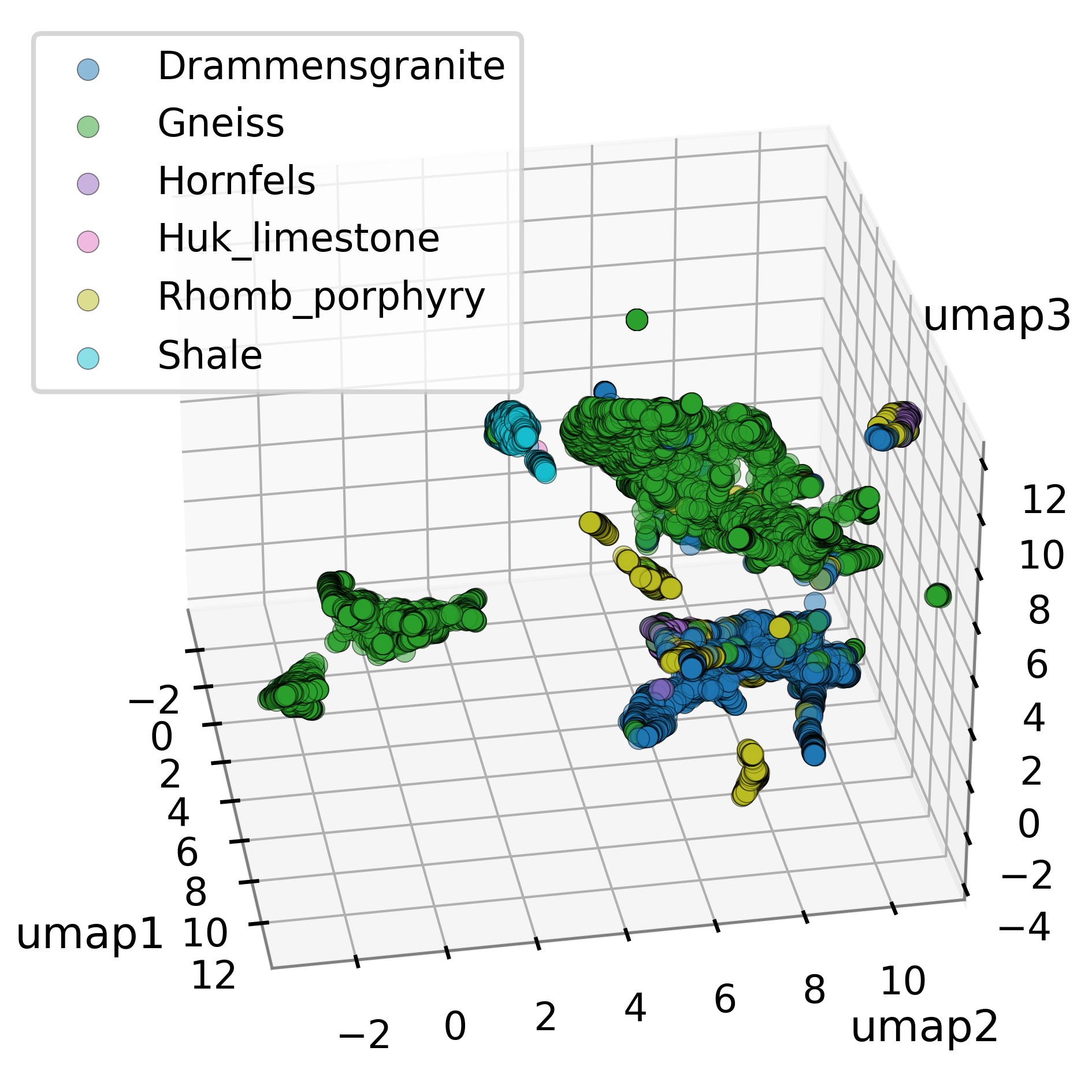} 
    \end{minipage}
    
    \caption{Experiment 7. Clustering of MWD data using UMAP for dimension reduction and HDBSCAN for clustering, \textbf{a}  Clusters coloured by cluster number, \textbf{b} Clusters coloured by rock quality (Q-class), \textbf{c} Clusters coloured by rock type. An interactive version of the plot is available in the digital version of the paper.}
    \label{fig:experiment7_mwd}
\end{figure}

The number and types of features significantly influence clustering outcomes. In the referenced plots, clustering quality appears nearly equal for `MWD' and `All', slightly better for the `All' set, which exhibits greater inter-cluster distance and cluster compactness. Including overburden thickness and tunnel width in the `All' set improves all performance metrics. This underscores the critical role of feature selection and the potential for cluster refinement by adding additional features to the MWD-feature base, characterising the rock mass. When adding new features, it is crucial to ensure a causal relationship based on domain knowledge relevant to the problem.

Transitioning from the 48-feature MWD set to the 30-feature MWD\_rock set generally results in slightly lower scores. The clustering quality visibly deteriorates, as evidenced by several clusters overlapping. Similarly, the MWD\_median set, with only eight features, shows acceptable scores but poor clustering on visual inspection, worse than the 30-feature set. The Gini index for the smaller feature sets indicates high values, suggesting the presence of one or two large clusters. This observation suggests that while high scores can correlate with effective clustering in 3D visualisations, they do not necessarily guarantee it; visual inspection remains essential to validate the results.

HDBSCAN, in contrast to agglomerative clustering, identifies samples as unclustered if they are outliers. Different experimental setups yield varying numbers of these unclustered samples, typically few. Fig.~\ref{fig:experiment7_mwd}a demonstrates the unclustered blue. Future research should investigate these outliers to potentially uncover patterns relevant to assessing rock mass stability.

\begin{figure}
    \centering
    \includegraphics[width=0.5\textwidth]{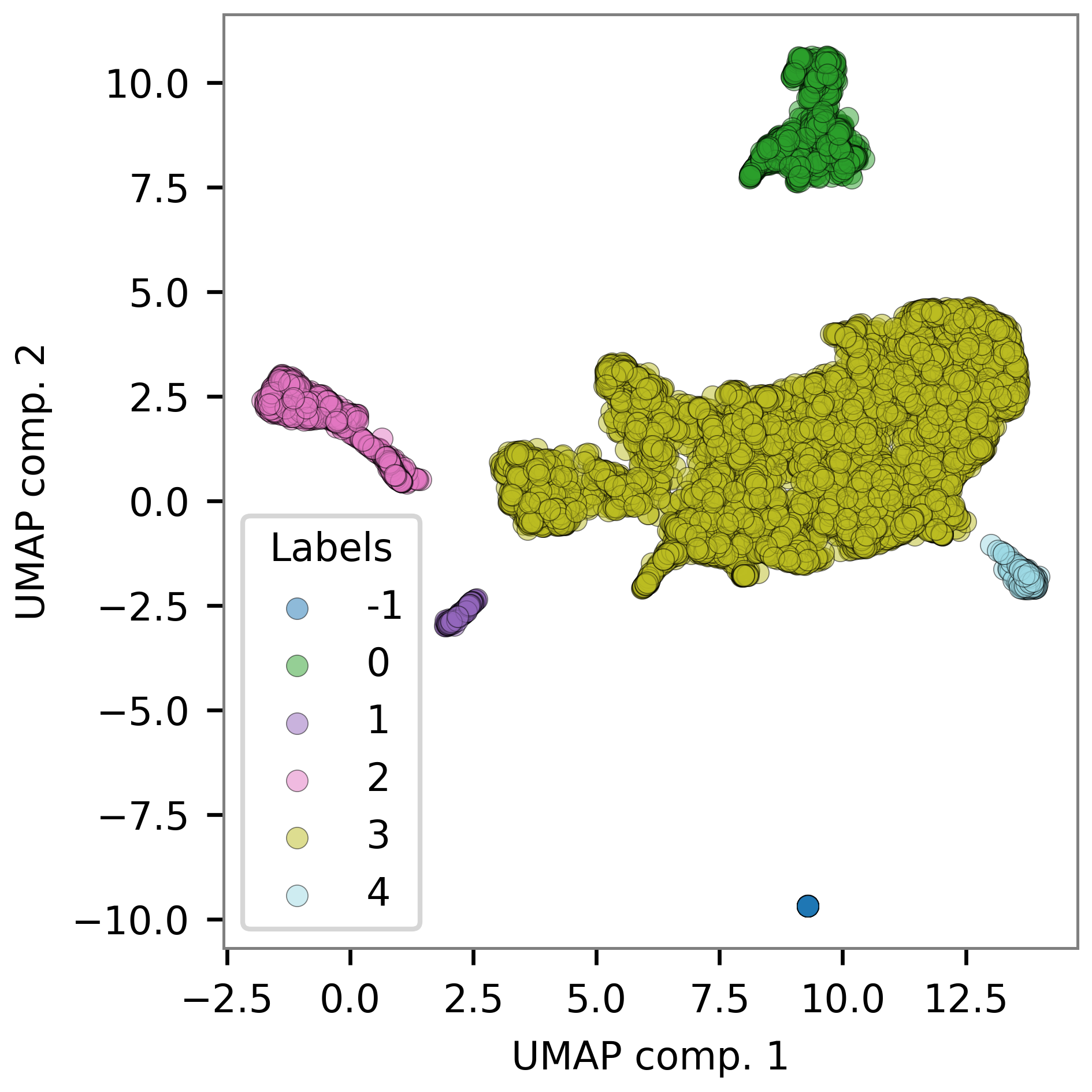}
    \caption{Experiment 10. Clustering of MWD data using UMAP for dimension reduction and HDBSCAN for clustering. Plot optimised for 2D view.}
    \label{fig:2Dscatter_cluster_2Dumap_hdbscan_mwd}
\end{figure}

\subsubsection*{Configurations leading to poor or questionable clustering}

Clustering results should not rely solely on the Silhouette score and Davies-Bouldin index; all three metrics are important. To illustrate this, we included experiment 12, which is notable for its lack of dimension reduction. Conducting 50 tests to optimise HDBSCAN without dimension reduction highlighted the necessity of this process for effective clustering. Despite yielding the best performance values, this experiment identified only three clusters, one encompassing over 99\% of the samples and two small ones. Similar outcomes were observed for K-means and agglomerative clustering without dimension reduction. Interestingly, experiment 10, with dimension reduction but only using two UMAP components, demonstrates some of the same patterns. Reducing UMAP components from three to two generally worsens the results, as shown in the poor default value results for experiments 0 and 7.

Including the Calinski-Harabasz index as a metric resulted in the exclusion of experiment 12 and also experiment 11, which utilised PCA for dimension reduction and HDBSCAN for clustering. Nonetheless, experiment 9, which combined PCA with K-means, also maintained a high Calinski-Harabasz score. Despite generally favourable metrics in the tables, K-means often underperform. This becomes apparent upon examining the plots, where multiple clusters overlap visually. The issue likely stems from K-means' preference for circular clusters, as noted in \citet{macqueen1967some}. This clustering method misrepresents rock mass data, which does not naturally form circular clusters, by forcing points into such clusters even when they do not logically fit. The only metric indicating this issue in Table~\ref{table:clustering_results} is the Gini index. K-means shows low values (below 0.3) for this index, suggesting nearly equal-sized clusters. However, it is unrealistic for rock mass data to naturally segment into ten clusters of equal sample size in each plot.

\subsection{Linking clusters to physical rock mass properties}\label{ordering_clusters}

In Table~\ref{tab:experiment0},~\ref{tab:experiment3}, and~\ref{tab:experiment7}, we present detailed results from experiments 0, 3, and 7. The clusters are arranged by increasing values of the Penetration MWD feature. It is important not to focus on the absolute values of the MWD features as they are scaled but rather on their relative sizes.

\begin{table}[h]
    \centering
    \caption{Experiment 0. Cluster properties for the feature set `all'. The label is rock quality (Q-class). The clustering algorithm is HDBSCAN. Values for three important cluster features are given to relate clusters to relatable properties. Clusters are ordered after increasing value for normalised penetration.}
    \label{tab:experiment0}
    \begin{tabular}{cccccc}
    \toprule
    cluster & Num & FeedPressNorm & PenetrNorm & Overburden & Label \\
            & samples & Median [bar] & Median [m/min] & [m] &  \\
    \midrule
    0 & 22 & 28.83 & -239.11 & 15.71 & E2 \\
    6 & 405 & -19.08 & -40.64 & 20.52 & B \\
    1 & 3680 & -1.22 & -9.67 & 170.77 & B \\
    7 & 3863 & 3.94 & -9.4 & 64.18 & B \\
    4 & 8445 & 1.08 & -6.12 & 63.51 & C \\
    8 & 1489 & -0.18 & -5.55 & 20.52 & C \\
    5 & 2169 & 18.77 & 13.57 & 34.71 & B \\
    3 & 2933 & -28.79 & 17.19 & 24.85 & C \\
    2 & 271 & -32.65 & 22.8 & 23.93 & C \\
    \bottomrule
    \end{tabular}
\end{table}

\begin{figure}[h]
    \centering 
    \includegraphics[width=0.9\textwidth,keepaspectratio]{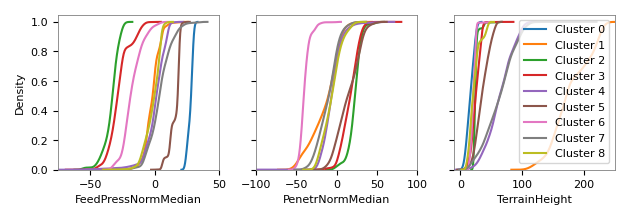} 
    \caption{Experiment 0. Cumulative distribution plots for three features separated in clusters.} 
    \label{fig:all_CDF} 
\end{figure}

Using the ' All ' feature set, we observe distinct patterns for experiment 0. The number of samples in each cluster is evenly spread, correlating with a low Gini score. Each cluster's median feature values are distinct, indicating differing physical properties. Had the median values been similar across multiple clusters, it would suggest poor clustering relative to rock mass properties. The overburden feature generally decreases as penetration increases, possibly due to reduced rock stress and more dayrock in areas with lower overburden and higher penetration. Cluster 0, containing only 22 samples, does not represent outliers but rather appears to reflect a niche geological phenomenon. Although small, the internal consistency of these samples suggests they share distinct properties that set them apart from larger clusters, indicating a rare but meaningful geological feature. 

The rock quality label from the Q-system is assigned to each cluster by majority vote. This is an example of the opposite of the good discrimination seen in the MWD-features. The Q-classes do not follow the clusters, as seen from the B and C-class samples spread out in all clusters. We can visually inspect the same bad alignment in Fig.~\ref{fig:experiment_0_all} and \ref{fig:experiment7_mwd}b and the low adjusted rand score for experiment 0 in Table~\ref{table:clustering_metrics}. This indicates that the natural clustering of the rock mass based on MWD features does not align with the Q-classes. The misalignment between classes in traditional rock mass classification systems and natural clusters from an automated, objective process highlights the limitations discussed in Section~\ref{limitations}. These include the non-uniqueness of existing values for specific rock masses and the challenge of reducing complex qualitative information to single numbers.  As pointed out in the last section, overburden thickness and tunnel width do not have any intuitive causal relationship with rock type. Still, the alignment with the rock type labels is pronounced in Fig.~\ref{fig:experiment_0_all}c, most likely caused by MWD data accounting for the major part of the feature vector for clustering. The cumulative distribution plots in Fig.~\ref{fig:all_CDF} illustrate a broad distribution for Feeder pressure. At the same time, penetration and overburden tend to form three groups, confirming the pattern observed in the median values analysis.

\begin{figure}[h]
    \centering 
    \includegraphics[width=0.9\textwidth,keepaspectratio]{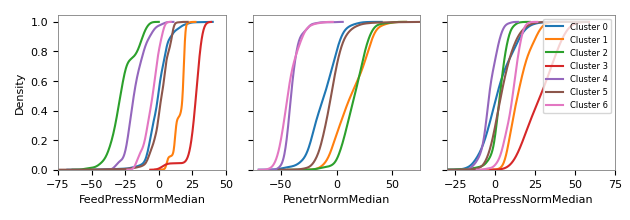} 
    \caption{Experiment 3. Cumulative distribution plots for three features separated in clusters.} 
    \label{fig:mwd_agglomerative_CDF} 
\end{figure}

For experiment 3, cluster 3 with only 23 samples have similarities with cluster 0 for experiment 0 in terms of its size and the values for feederpressure and penetration rate. The median values are also here clearly separable for each cluster (seen from Table~\ref{tab:experiment3} and CDF's in Fig.~\ref{fig:mwd_agglomerative_CDF}), with a tendency of three groups for penetration and a well spread of values for feeder pressure and rotation pressure (torque). Rock types have been assigned to each cluster by majority vote. An adjusted rand score of 0.3 indicates a reasonable alignment with the rock type labels. The lowest penetration value for the strong Hornfels rock type and the highest penetration for the weaker Rhomb porphyric rock align well with the physical world.

\begin{table}[h]
    \centering
    \caption{Experiment 3. Cluster properties for the feature set `mwd'. The label is rock type. The clustering algorithm is Agglomerative clustering. Dimension reduction is UMAP. Values for three cluster features are given to relate clusters to relatable properties. Clusters are ordered after increasing value for normalised penetration.}
    \label{tab:experiment3}
    \begin{tabular}{@{}cccccc@{}}
    \toprule
    cluster & Num & FeedPressNorm & PenetrNorm & RotaPressNorm & Label \\
            & samples & Median [bar] & Median [m/min] & Median [bar] &  \\
    \midrule
    3 & 23 & 28.4 & -239.08 & 29.47 & Granittic\_gneiss \\
    6 & 493 & -4.25 & -43.78 & 11.16 & Hornfels \\
    4 & 422 & -18.67 & -40.67 & -4.03 & Rhomb\_porphyry \\
    0 & 7477 & -0.31 & -10.93 & 0.88 & Drammensgranite \\
    5 & 8942 & 2.19 & -4.83 & 3.61 & Granittic\_gneiss \\
    1 & 2365 & 18.25 & 12.42 & 14.04 & Granittic\_gneiss \\
    2 & 1823 & -28.52 & 16.44 & 2.91 & Rhomb\_porphyry \\
    \bottomrule
    \end{tabular}
    \end{table}

\begin{figure}[h]
    \centering 
    \includegraphics[width=0.9\textwidth,keepaspectratio]{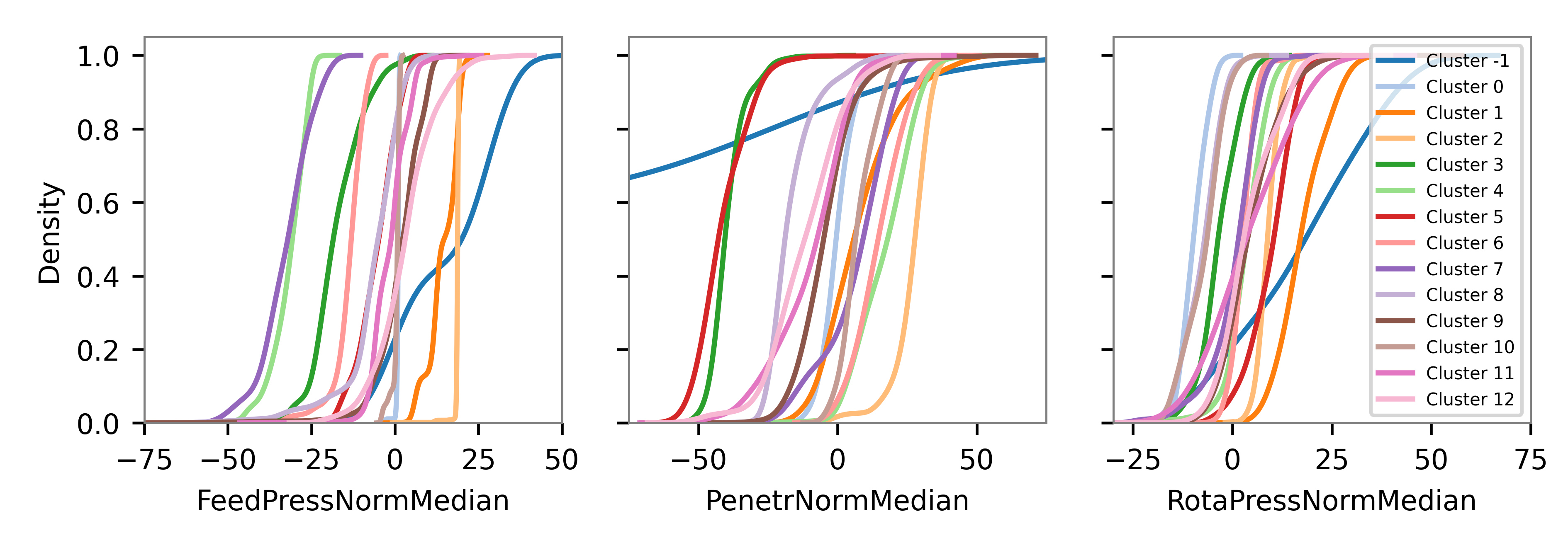} 
    \caption{Experiment 7. Cumulative distribution plots for three features separated in clusters.} 
    \label{fig:mwd_hdbscan_CDF} 
\end{figure}

Twenty-three samples are not classified in experiment seven (see Table~\ref{tab:experiment7}). These are visible as blue-coloured points in the scatter plot of clusters in Fig.~\ref{fig:experiment7_mwd} and highlighted by the blue-coloured outlier distribution in for the CDF plots of Feederpressure, Penetration and Rotationpressure in Fig.~\ref{fig:mwd_hdbscan_CDF}. What is apparent in experiment 7, compared to the other experiments, is the increased separation of cluster properties seen by all the Median values for MWD-features in Table~\ref{tab:experiment7} and CDFs in Fig.~\ref{fig:mwd_hdbscan_CDF}. This indicates a clear separation of rock mass properties in the twelve defined clusters. We have assigned rock type labels to the clusters by majority vote. In Fig.~\ref{fig:experiment7_mwd}c, we have coloured the scatter points with rock type labels, clearly illustrating the alignment of rock type to the clusters. This alignment is confirmed by the highest adjusted Rand score of the experiments of 0.47 (a score of 1.0 would be a perfect match between clusters and rock type).  Such an alignment is also confirmed by the study of \citet{Hansen2024rocktype}, which forecasted rock type from MWD-data with high predictive accuracy (above 96\%).

\begin{table}[h]
    \centering
    \caption{Experiment 7. Cluster properties for the feature set `MWD'. The label is rock type. The clustering algorithm is HDBSCAN. Dimension reduction is UMAP. Values for three cluster features are given to relate clusters to relatable properties. Clusters are ordered after increasing value for normalised penetration.}
    \label{tab:experiment7}
    \begin{tabular}{@{}cccccc@{}}
    \toprule
    cluster & Num & FeedPressNorm & PenetrNorm & RotaPressNorm & Label \\
            & samples & Median [bar] & Median [m/min] & Median [bar] &  \\
    \midrule
    -1 & 38 & 24.98 & -237.93 & 17.81 & Granittic\_gneiss \\
    5 & 533 & -4.53 & -43.1 & 10.56 & Hornfels \\
    3 & 416 & -18.67 & -40.53 & -3.89 & Rhomb\_porphyry \\
    8 & 1141 & -5.16 & -19.18 & -6.92 & Granittic\_gneiss \\
    12 & 2917 & 3.58 & -11.72 & 2.69 & Drammensgranite \\
    11 & 3316 & -0.44 & -7.18 & 3.35 & Drammensgranite \\
    9 & 8720 & 2.72 & -4.95 & 3.79 & Granittic\_gneiss \\
    0 & 91 & 1.08 & -0.88 & -10.0 & Granittic\_gneiss \\
    1 & 1700 & 14.96 & 5.74 & 16.99 & Granittic\_gneiss \\
    10 & 147 & 0.84 & 6.08 & -6.04 & Granittic\_gneiss \\
    7 & 178 & -31.07 & 9.18 & 1.63 & Rhomb\_porphyry \\
    6 & 481 & -13.08 & 14.19 & 2.62 & Rhomb\_porphyry \\
    4 & 1085 & -30.61 & 18.19 & 3.49 & Blackshale \\
    2 & 698 & 18.77 & 27.76 & 8.8 & Augen\_gneiss \\
    \bottomrule
    \end{tabular}
\end{table}

As for experiment 3, the strong Hornfels are correctly assigned to a cluster (no. 5) with the lowest penetration and the next-to-highest torque. Regarding penetration, we have an overall pattern for big clusters going from Hornfels to Granite, Gneiss and Shale, which seems like an intuitive order. Rhombphorphyry is a lava rock with distinct properties for different lava flows. The samples of Rhombporfhyry are from several lava flows. That fact can explain why Rhombphorphyry has both low and high penetration. Augen Gneiss (cluster no. 2) is a special kind of Gneiss with specific properties, with its eye-shaped mega crystals of feldspar in a matrix of quartz and other minerals, quite different from the other classicly layered Gneisses in the other samples. Evaluation only based on penetration is too simple for this strong rock type. Inspecting Feeder pressure (note the pronounced CDF-distribution in Fig.~\ref{fig:mwd_hdbscan_CDF}) and Rotation pressure, we see high values, reflecting the strongness of this rock type. We, therefore, can explain its label on cluster number two.

\section{Discussion\label{discussion}}
In this section, we \textbf{a} outline a generic process for designing rock mass classification systems aimed at different objectives (e.g. stability risk assessment, rock support requirement, grouting effort, blastability) using the established clusters as a foundation, \textbf{b} explore a promising system approach, and \textbf{c} detail the applications of the established concept.

When designing the system using the process outlined in the next Section~\ref{sec:sketching_system}, we primarily adhere to the purpose and requirements for a rock mass classification system described by \citet{Bieniawski1973}. According to \citet{Bieniawski1973}, a rock mass classification should:

\textit{\begin{enumerate}
    \item Divide a rock mass into groups of similar behaviour;
    \item Provide a good basis for understanding the characteristics of the rock mass;
    \item Facilitate the planning and design of structures in rock by yielding quantitative data required for the solution of real engineering problems;
    \item Provide a common basis for effective communication among all persons concerned with a geomechanics problem.
\end{enumerate}
}
However, we consider rock mass classification in a broader sense, with different end goals compared to existing systems. We focus on using a core signature of the rock mass, largely independent of existing forces acting on the material, which contrasts with the traditional approach of enforcing similar behaviour. This core signature represents the rock mass as defined in ISO 14689\citep{iso14689}—\textit{rock comprising the intact material together with the discontinuities and weathering zones}. The system can then be tuned to specialised use cases, such as assessing advance rock support.

\subsection{Towards a generic data-driven rock mass classification system\label{sec:sketching_system}}

\begin{figure}
    \centering
    \includegraphics[width=1\linewidth]{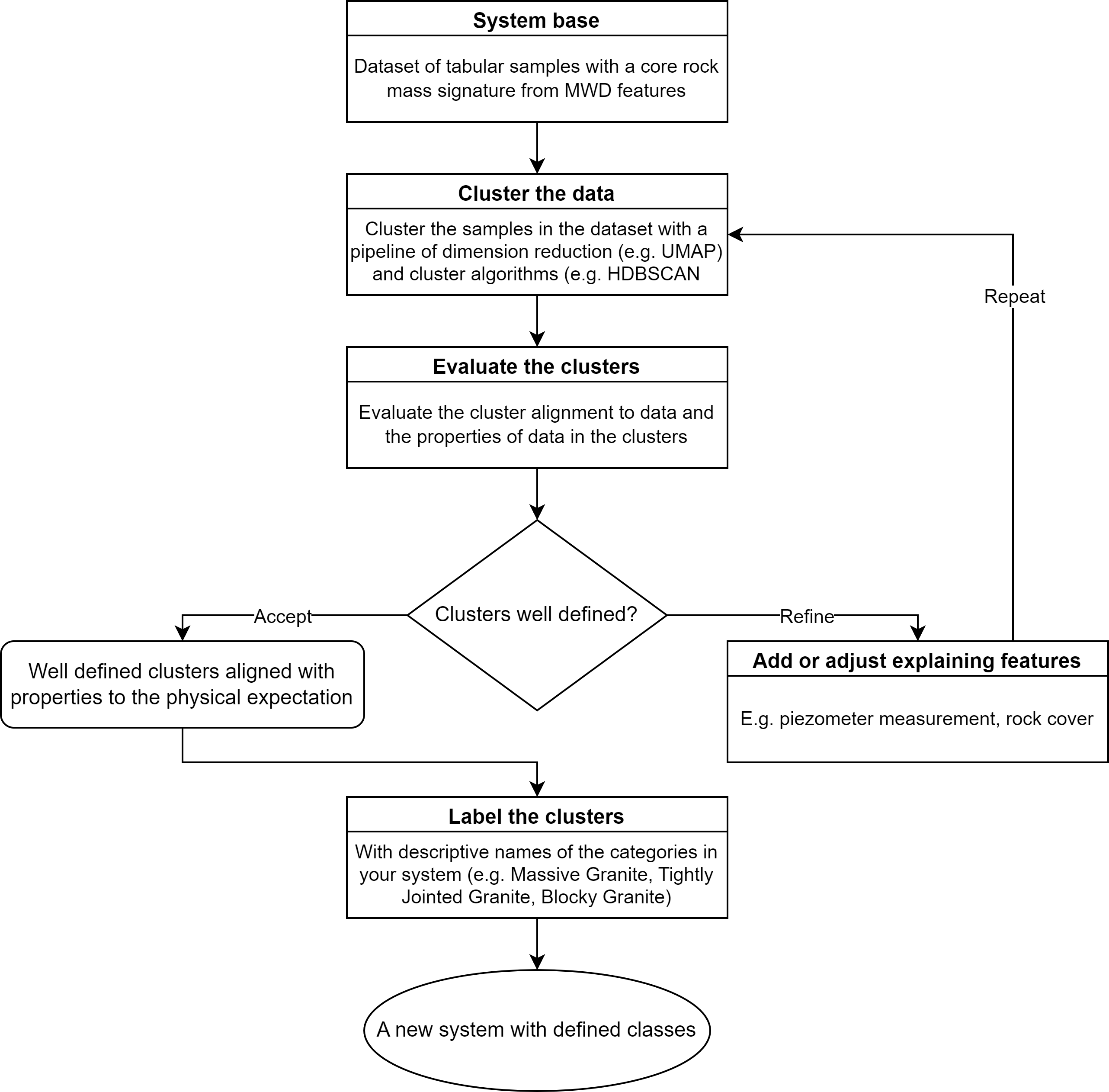}
    \caption{Flowchart for designing a cluster-based classification system}
    \label{fig:system_flowchart}
\end{figure}

MWD-data, represented as extracted vectors of statistical metrics from thousands of values, can be clustered into well-defined clusters. To develop a rock mass classification system, it is essential to thoroughly examine and understand the characteristics of each cluster in relation to the specific problem. The arrangement of classes may vary from weak to strong, low risk to high risk, or something structurally different, depending on the problem requirements. To ensure the clusters are relevant and achieve your purpose, you must \textit{tune the clusters to your problem} to exhibit desired properties. The resulting clusters (size, numbers, shape) will vary depending on the selected features. To obtain meaningful, well-defined clusters fitted to your problem, we suggest following the steps below, schematically illustrated in Fig.~\ref{fig:system_flowchart}. The process is intended to be implemented programmatically using a programming language such as Python (see Section~\ref{experimentation_process} and supplementary material in Section~\ref{sec:supplementary_material} for more details).

\textbf{System base}. Base your system on a core dataset of MWD features, such as the 48-value feature vector described in this study, which serves as a signature for the rock mass and forms somewhat defined clusters. Later, you will refine these clusters by adding new domain-specific features. 

\textbf{Cluster the data} using a pipeline of dimension reduction and clustering algorithms outlined in this study. Optimise parameters to algorithms if necessary

\textbf{Evaluate the clusters} by first examining their alignment with the data in 3D plots and assessing cluster performance metrics. Next, investigate the cluster properties by analysing the distributions of cluster features for physical alignment with the problem. Start with the distribution of features already included. Continue with existing logged values yet to be added as features, such as the number of joint sets or values from point load testing. Other more qualitative analyses may also be conducted to check the alignment. Subsequently, gather additional data from the sample sites in the dataset. This property evaluation of the clusters is crucial. If the data properties of the clusters do not show expected characteristics, such as defined differences in UCS values or joint infill between clusters, the clustering is misaligned with the problem, and you need to add or adjust the features used in clustering.

\textbf{Refine the clusters by adding features}. Adapt these clusters to your problem through further refinement, creating more defined clusters and, eventually, subclusters. Incorporate important features for the problem, ensuring they are easily and consistently measurable at the excavation site. Potential features include rock cover, tunnel width, measured water inflow in drillholes, distance to a parallel tunnel, soil cover, distance to a fault, point load value, and distance to a lake. By adding the easily collectable features of rock cover and tunnel width to the core MWD clusters in this study, we demonstrated a significant enhancement of the clustering in both visual and metric terms.

Further refinement may require additional harder-to-collect features to achieve a sufficiently detailed cluster system, possibly to accommodate different failure mechanisms. Given their significant impact on expected failure types, the principal rock stress components are likely essential in some systems. Unlike the time-consuming experiments needed to measure rock mechanical properties (E-modulus, UCS, Poisson's ratio), which also face scaling issues, approximating rock stress might be feasible and sufficiently accurate for considering rock failure principles. The addition of overburden thickness, a critical variable in estimating vertical stress, exemplifies such a tuning step (see Section~\ref{comparing_clustering}).

When clustering your data, you are not constrained by the requirements of supervised learning, where the model must generalise based on features present in both training and prediction phases. This allows the use of any features in your dataset to form clusters, even those not available during the prediction of new samples. For example, post-blasting metrics like overbreak can be included in clustering but are unsuitable for supervised learning. However, as described below, you cannot use these post-excavation features when training a supervised learning model to classify new samples. 

Repeat the refinement process until the clusters align with your problem regarding properties, count, and definition. Finally, label the clusters with descriptive names that reflect the categories in your new classification system.

\textbf{Action mapping}. When the classification system is established, you need to map actions to the clusters, such as associating support classes with RMR and the Q-system or grouting effort described by the number of grouting holes and the volume of grouting material. One approach would be to map empirical data of rock support from stable sites to the clustering value of that site (e.g. complementary to support classes for stability classes A, B, C\dots in the Q-system). Alternatively, you could incorporate descriptions of installed support from stable sites into the cluster feature. This approach would allow clusters to encompass information about specific rock masses and the corresponding stable rock support. When a new rock mass is encountered during excavation, the appropriate rock support can be directly inferred from the cluster. Other methods may also be viable, but exploring these is beyond the scope of this study.

\textbf{Supervised learning model for future prediction}. To implement the clustering-based system, the standard approach is to build a supervised learning model to classify new rock masses encountered during excavation into the correct cluster label. \citet{Sapronova2021sparse} outlines a relevant approach. The samples used for clustering are annotated with the new cluster labels, and a supervised model is then trained on these samples, following a systematic procedure such as the one outlined in \citet{Hansen2024rocktype}. 

A supervised learning model in prediction mode facilitates the assessment of misclassification probability, providing an uncertainty measure for classification. This is an important requirement for a rock mass classification system, as highlighted by \citet{stille_classification_2003}. If the classification process identifies a rock mass as an outlier, the new sample should be closely inspected. If no anomalies are found, it suggests that the rock mass type is not included in the existing system, indicating a need for system updates through re-clustering.

While it is possible to use the fitted clustering model directly to label new data, training a supervised learning model on the clustering results offers several advantages. These benefits include improved prediction accuracy, the ability to incorporate new features, enhanced assessment of prediction metrics and uncertainty, increased flexibility in hyperparameter tuning, and the use of techniques such as cross-validation to improve generalisability to new data.

When encountering a rock mass type or characteristics not previously seen, the fitted HDBSCAN clustering model can label such a sample as noise (i.e., an outlier) if it does not belong to any of the existing clusters based on density criteria. This flexibility allows the system to handle novel or atypical conditions without forcing the sample into an existing cluster, ensuring that truly new geological phenomena are appropriately flagged for further analysis.

To reduce the frequency of new samples being labelled as outliers, it is important to fit a clustering model and train a supervised model on a large dataset with a diverse range of rock mass types. By doing so, the models can form a more representative and generalisable understanding of various geological conditions, thus reducing the need for frequent re-modelling. The present study is intended to demonstrate the concept, but for practical deployment in production, using a broad range of rock masses is essential to ensure robustness and minimise the occurrence of outliers.

Using this process, a data-driven classification system can be developed \textit{tailored to a specific problem or site}. This study focuses on rock mass stability, primarily concerning permanent or advance rock support. Alternatively, another system could concentrate on blasting design (geometry, number of holes, specific charge), grouting effort (grouting volume, pumping time), or water leakage. The selection of features and the design of clusters would be made accordingly, to best suit the intended application. Following the successful alignment of rock type to clusters, as detailed in experiment 7 in Section~\ref{ordering_clusters}, we describe a concrete approach below to initiate the design of an intuitive classification system.

\subsection{A promising classification approach} 

The significant alignment between rock types and clusters suggests the potential for developing an intuitive rock mass classification system.  Ground control engineers intuitively understand and accept the differences in required rock support between rock types, such as blocky versus tightly jointed Granite or a brittle, porous Rhombphorphyry versus a massive one. This understanding can extend to expectations of water problems or drill bit wear. By labelling the clusters (i.e. the rock mass classes) with descriptive names like "Massive Granite", "Blocky Granite", and "Tightly Jointed Granite", it becomes easier to determine the appropriate rock support for each class. Although this approach may result in a large number of classes, such descriptive labels are likely more intuitive and meaningful for engineers than existing alphanumeric systems. An engineer knows the implications of a blocky Granite in terms of rock support, water leakage, etc., but may not fully understand what a Q-class C rock mass entails. However, a clearer distinction between different rock types (e.g. as seen from Shale and Limestone majorly falling into the same cluster in Fig.~\ref{fig:appendix_mwd_experiment_7_detailed_geology} in Appendix~\ref{appendix:section_cluster_performance}) with joints and weathering, are needed before such a system can be deployed. Features that could improve clustering include the accurate measurement of joint sets, point load testing, and a categorical feature that describes the appearance of joint material.

\subsection{Applicability and reproducibility considerations}

In this study, we have clustered MWD-data, framed as vectors of extracted statistical metrics, collected from thousands of values in one-meter sections of a full-face blasting round. Conceptually, the approach might be used on all kinds of MWD-data, from drilling boltholes, exploratory holes, grouting, and blasting holes, provided a sufficient quantity of MWD-values to extract the statistical vector that characterises the rock mass. For instance, the vector could be computed for $1\times1\times1$ meter cubes of data, $4\times4\times4$ meter cubes, a 1-meter long split profile section, or 1-meter sections in single holes. Determining the limits of this application is a subject for future research.

To reproduce the results, including pronounced clustering and comparable metrics, using the same methodology for another dataset, it is advisable to use MWD data from at least three significantly different rock types. This diversity leads to distinct MWD signatures, enabling the identification of more than two clusters and providing a clear sense of the clustering potential. The method with the best performance in this study, HDBSCAN, is generally considered robust to sample sizes, but its performance improves with larger sample sizes in high-dimensional data. Literature suggests that around 100 to 1000 samples per cluster are sufficient, resulting in a total dataset of a few thousand samples depending on the number of classes \citep{Sarstedt2019, atif2024least}.

\subsection{Limitations and Cautions in Applying Machine Learning to Rock Engineering }
While applying machine learning (ML) in rock engineering can enhance the accuracy and objectivity of rock mass classification, it is crucial to recognise its limitations and the caution required in its implementation. Unlike traditional methods relying on expert judgment and empirical correlations, ML approaches introduce challenges such as the need for large, high-quality datasets and the risk of overfitting or misinterpreting results due to complex model structures. 

A classification system based on data-driven machine learning algorithms may be perceived as less transparent than existing systems, making it difficult for practitioners to understand or trust these algorithms' decisions fully. However, compared to the subjective judgement and unspecified design processes in existing systems (ref. point~\ref{item:update} in Section~\ref{limitations}), the machine learning-based process can be largely deterministic, mathematical, and reproducible, such as the methods described in this study. Although challenges of transparency and reproducibility remain for certain models, particularly neural network-based models, there is an increasing focus on the explainability and interpretability of machine learning models. For supervised learning models, well-proven methods explain classifications in terms of the impact of each feature on the outcome\citep{Hansen2024_explainability}. By choosing models wisely, it is possible to interpret the inner workings of the model \citep{Hansen2024_explainability}, and this is increasingly true for unsupervised learning models as well \citep{ALVAREZGARCIA2024120282_explainable}.

An ML model is not a silver bullet; while it can reduce subjectivity and improve predictive power, it must be used as decision support alongside domain expertise and traditional approaches. Future research should focus on developing a combined approach that increasingly integrates ML models with established rock engineering practices, ensuring the advantages of both approaches are fully leveraged while minimising potential risks.

\section{Conclusions}\label{conclusion}

Rock mass classification systems are crucial for mapping risks and guiding support and excavation design globally. However, systems developed primarily in the 1970s lack access to modern high-resolution datasets and advanced statistical learning techniques, which limits their effectiveness as decision-support systems. We have demonstrated that a pipeline of dimension reduction and unsupervised machine learning can effectively form well-defined clusters using extracted statistical information from thousands of MWD-data values representing the whole encountered rock mass for 1-meter sections in infrastructure tunnels. Such clusters can serve as a robust foundation for various rock mass classification systems. The study yields the following conclusions:

\begin{itemize}
    \item A representation learning pipeline consisting of a min-max scaler, dimensionality reduction with UMAP, and clustering using HDBSCAN or Agglomerative Clustering has proven effective for clustering rock mass data. This effectiveness is evident both visually, in a 3D scatter plot of three UMAP components, and numerically, as demonstrated by various cluster evaluation metrics. 
    \item Clustering efficiency depends on the number and type of features. Optimal results were achieved with the largest set of 50 features, which included two geometric features. The set of 48 values, including only MWD features, also showed effective clustering. However, the 50-value set is not optimal in an absolute sense and cannot be directly compared to the 48-value set. Including additional features to the core MWD set was demonstrated to refine and improve clustering performance. Among MWD feature sets, the maximum set of 48 values performed significantly better than smaller sets.
    \item Multi-objective optimisation of algorithm parameters was crucial for achieving effective clustering. Over 1000 experiments were conducted, with most failing. 
    \item Evaluation of clustering effectiveness required rigorous assessment using internal cluster metrics (Silhouette Coefficient Score, Davies Bouldain Index, Calinski Harabasz Index), the Adjusted Rand Score, and the Gini-index, alongside the count of clusters and unclustered samples, complemented by a visual review in a 3D interactive plot. 
    \item The Gini index may serve as a final evaluator when other scores are inconclusive, albeit with nuances. We discarded pipelines with K-means due to their low Gini-index (all other scores were favourable), indicating unnaturally even cluster sizes, coupled with a visual inspection. Several experiments were also discarded due to excessively high Gini-index values, typically indicating one dominant cluster and a few smaller ones. 
    \item Effective clustering necessitates dimension reduction. The non-linear UMAP algorithm was successfully employed, whereas clustering without dimension reduction or using the linear PCA algorithm yielded poor results.
    \item Concerns regarding the use of a manifold learning technique like UMAP, due to its tendency to occasionally create artificial clusters without physical significance, were disproved. Clear, physically meaningful clusters were demonstrated, contrasting with the ineffectiveness of no dimension reduction or the linear PCA method. 
    \item Of the two label sets available, the rock type aligned well with the defined clusters. The rock mass quality labels (Q-class) aligned poorly, with labelled samples randomly spread out in the clusters.
    \item Clusters with a core of MWD-features, which act as a signature of the encountered rock mass, can be refined with domain-specific features like rock cover and tunnel width. 
    \item Analysis of feature distributions involved in clustering revealed that the physical properties of each cluster are specific and align well with a range from weak to strong rock types, indicating meaningful clustering. 
\end{itemize}

\section{Outlook and future research}\label{outlook}

The successful clustering of MWD-data into defined groups, the distinct physical properties of each group, the demonstrated feature tuning possibilities, and the alignment of these clusters with rock types suggest that the described methods could be a foundational basis for new data-driven rock mass classification systems, such as stability systems. Rather than developing a universal stability classification system that encompasses all variants—such as different failure systems (squeezing, low stress, swelling), rock mass categories (strong elastic-behaving rocks, weak plastic rocks), and tunnel geometries—it may be more effective to tailor systems to specific problems and sites. This could be achieved by adapting the classification to the rock mass signature derived from MWD values and incorporating additional features relevant to the problem. The systems can be named accordingly, such as "MWD-system-hard-rock" or "MWD-system-squeeze". 

Key areas for future research to further develop a data-driven classification system for rock mass stability include: (a1) closely examining the properties of each rock mass cluster, (a2) refining the clusters by incorporating relevant domain features to address specific problems, (a3) ensuring the clusters are organised meaningfully and align with the particular problem, and labelling them with appropriate stability names, (b) mapping actions to these clusters, and (c) training a supervised learning model to accurately classify new rock masses into the appropriate clusters.

Further research is motivated by the potential of such a system to address the limitations of existing classification systems, as outlined in the introduction Section~\ref{limitations}. A data-driven approach based on MWD-data is likely to overcome several of these limitations:

\begin{itemize}
    \item Automated data collection from sensors provides high-resolution coverage of the entire rock mass without the need for human data assessment, addressing limitations such as~\ref{item:human_bias}-Human bias, \ref{item:perception}-Inconsistent-assessment, \ref{item:hazardous_inspection}-Hazardous-inspection, and \ref{item:quantification}-Non-representative-quantification.
    \item Conceptually, the MWD-data signature can be established for small volumes, as long as there are enough MWD-values to extract a signature, say for a $1\times1\times1$ $m^3$ resolution, allowing for fine-grained assessment and targeted rock support, overcoming limitation \ref{item:not_finegrained}-Not-finegrained-support.
    \item Expanding the dataset with new features and site samples (e.g., different geological conditions or tunnel uses) and retraining the cluster models is an efficient and transparent process, addressing limitation \ref{item:empirical}-Limited-case-studies and \ref{item:update}-Cumbersome-update. 
    \item Adapting the system for specific cases, such as mine junctions and tunnel openings, might be managed by adding predictive causal features to the dataset and rerunning the clustering, overcoming the limitation \ref{item:complexity}-Complex-exception-rules.
    \item MWD-data for clustering is efficiently collected from all drillholes, from boltholes radially around the tunnel profile, from blasting holes ahead, and from exploratory holes ahead and outside of the profile, overcoming limitation \ref{item:visual_asessment}-Only-visual-assessment and \ref{item:advance_support}-No-advance-assessment.
    \item A clustering approach simplifies the process of defining multiple classes, potentially allowing for more homogeneous rock mass material classification and targeted rock support with specific safety factors, avoiding a conservative approach with few classes, overcoming limitation \ref{item:concervative_support}-Non-optimised-rock-support and \ref{item:right_support}-Correct-rock-support.
    \item The flexibility to define multiple classes, adding causal features, and tune classes to specific properties may also address limitation \ref{item:failure_modes}-Not-representative-failure-modes.
\end{itemize}

MWD is cost-effective and readily available in tunnelling and mining worldwide, making it a viable foundation for rock mass classification systems. The findings of this study require further validation, as summarised in this section and detailed in Sections~\ref{results} and~\ref{discussion}, particularly regarding the properties and structures of the established clusters. If confirmed, the implications for tunnelling and mining could be substantial. The same form of representation learning outlined in this study should likely be possible to apply to detect classification patterns in the rock mass using other forms of high-resolution datasets, such as face geophysics or lidar scanning of exposed rock, thereby increasing the method's impact.

Following the procedures outlined in this study, which utilise core clusters based on rock mass signatures from MWD-data, it should be feasible to develop purely data-driven rock mass classification systems with specific targets like stability, grouting effort, or blastability. Predicting cluster labels from data obtained from long exploratory holes would enable rock mass classification several days in advance, enhancing planning capabilities. Such a decision support system would potentially be easily updatable, transparent, reproducible, and free from human bias. Optimised to address the limitations of existing systems, it could significantly impact the industry and society by optimising decisions, reducing the use of steel and concrete, enhancing safety by mitigating risks associated with complex geology, and increasing tunnelling efficiency.

\section{Acknowledgement}

The authors gratefully acknowledge the tunnel software/hardware company Bever Control, which has facilitated the data from the clients Bane NOR, Statens Vegvesen, Nye Veier, and the contractor AF-Gruppen.

\section{Ethics declarations}

\textbf{\large Conflict of interest}\\
The authors declare that they have no known competing financial interests or personal relationships that could have appeared to influence the work reported in this paper.\\ 

\noindent\textbf{\large Funding}\\
This research received no specific grant from funding agencies in the public, commercial, or not-for-profit sectors.\\

\noindent\textbf{\large Contributions}\\
Tom F. Hansen: Conceptualisation, Methodology, Software, Investigation, Data Curation, Visualisation, Writing --- Original Draft. Arnstein Aarset: Conceptualisation, Writing --- Review \& Editing. \\

\noindent\textbf{\large The use of generative AI in the writing process}\\
While preparing this work, the authors used GPT-4 from OpenAI to improve the readability and language of some paragraphs in the text. After using this tool/service, the authors reviewed and edited the content as needed, and take full responsibility for the content of the publication.

\section{Supplementary material\label{sec:supplementary_material}}

The code for the project is available under the following Github Repository: 
\url{https://github.com/tfha/ML-MWD-clustring}

\bibliography{references_revision1.bib}

\clearpage

\appendix
\appendixpage

\section{Hyperparameter details}\label{appendix:section_hyperparameters}

\begin{table}[!h]
    \centering
    \caption{Best parameters from hyperparameter optimization for three different pipelines including UMAP and a clustering algorithm. Default value is given in parenthesis}
    \label{appendix:hyperparameters}
    \begin{tabular}{lll}
    \toprule
    Algorithm & Parameter & Value \\
    \midrule
    \multicolumn{3}{c}{Experiment 0} \\
    \midrule
    HDBSCAN & min\_cluster\_size & 22 (5)\\
            & min\_samples & 13 (None)\\
            & metric & chebyshev (euclidean)\\
            & cluster\_selection\_epsilon & 0.340 (0.0)\\
    UMAP    & n\_neighbors & 197 (15)\\
            & min\_dist & 0.0 (0.1)\\
            & n\_components & 12 (2)\\
            & metric & euclidean (euclidean)\\
    
    \midrule
    \multicolumn{3}{c}{Experiment 3} \\
    \midrule
    Agglomerative Clustering & n\_clusters & 7 (2)\\
                             & metric & cosine (euclidean)\\
                             & linkage & average (ward)\\
                             & distance\_threshold & None (None)\\
    UMAP                     & n\_neighbors & 46 (15)\\
                             & min\_dist & 0.0 (0.1)\\
                             & n\_components & 6 (2)\\
                             & metric & euclidean (euclidean)\\
    
    \midrule
    \multicolumn{3}{c}{Experiment 7} \\
    \midrule
    HDBSCAN & min\_cluster\_size & 83 (5)\\
            & min\_samples & 14 (None)\\
            & metric & manhattan (Euclidean)\\
            & cluster\_selection\_epsilon & 0.690 (0.0)\\
    UMAP    & n\_neighbors & 21 (15)\\
            & min\_dist & 0.0 (0.1)\\
            & n\_components & 3 (2)\\
            & metric & euclidean (euclidean)\\
    \bottomrule
    \end{tabular}
\end{table}

\clearpage

\section{Detailed cluster performance figures}\label{appendix:section_cluster_performance}

\begin{figure}[!h]
    \centering 
    \includegraphics[width=1.0\textwidth,keepaspectratio]{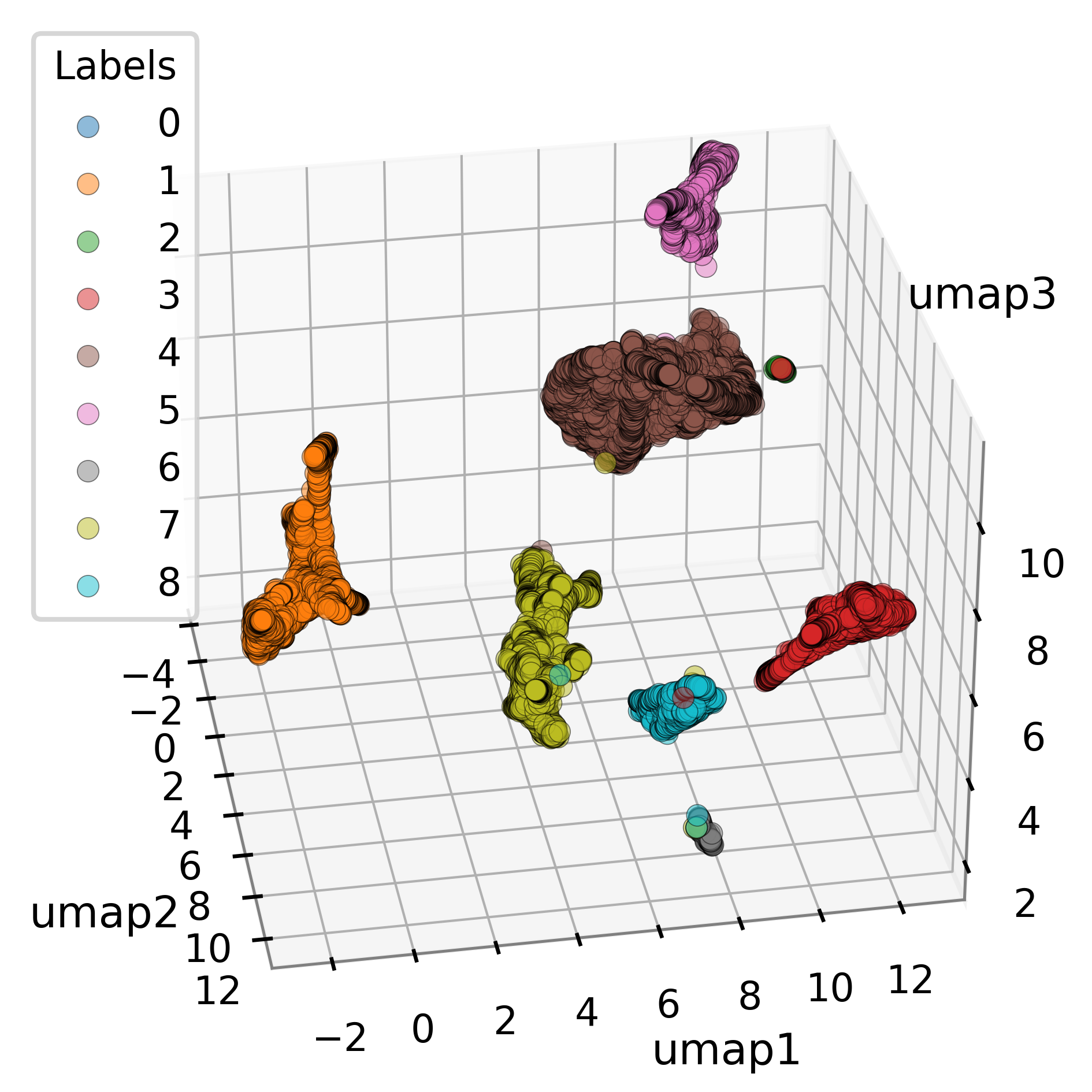} 
    \caption{Coloured clusters for experiment 0, on featureset `all' using HDBSCAN for clustering and UMAP for dimension reduction.} 
    \label{fig:appendix_all_3D} 
\end{figure}

\begin{figure}[!h]
    \centering 
    \includegraphics[width=1.0\textwidth,keepaspectratio]{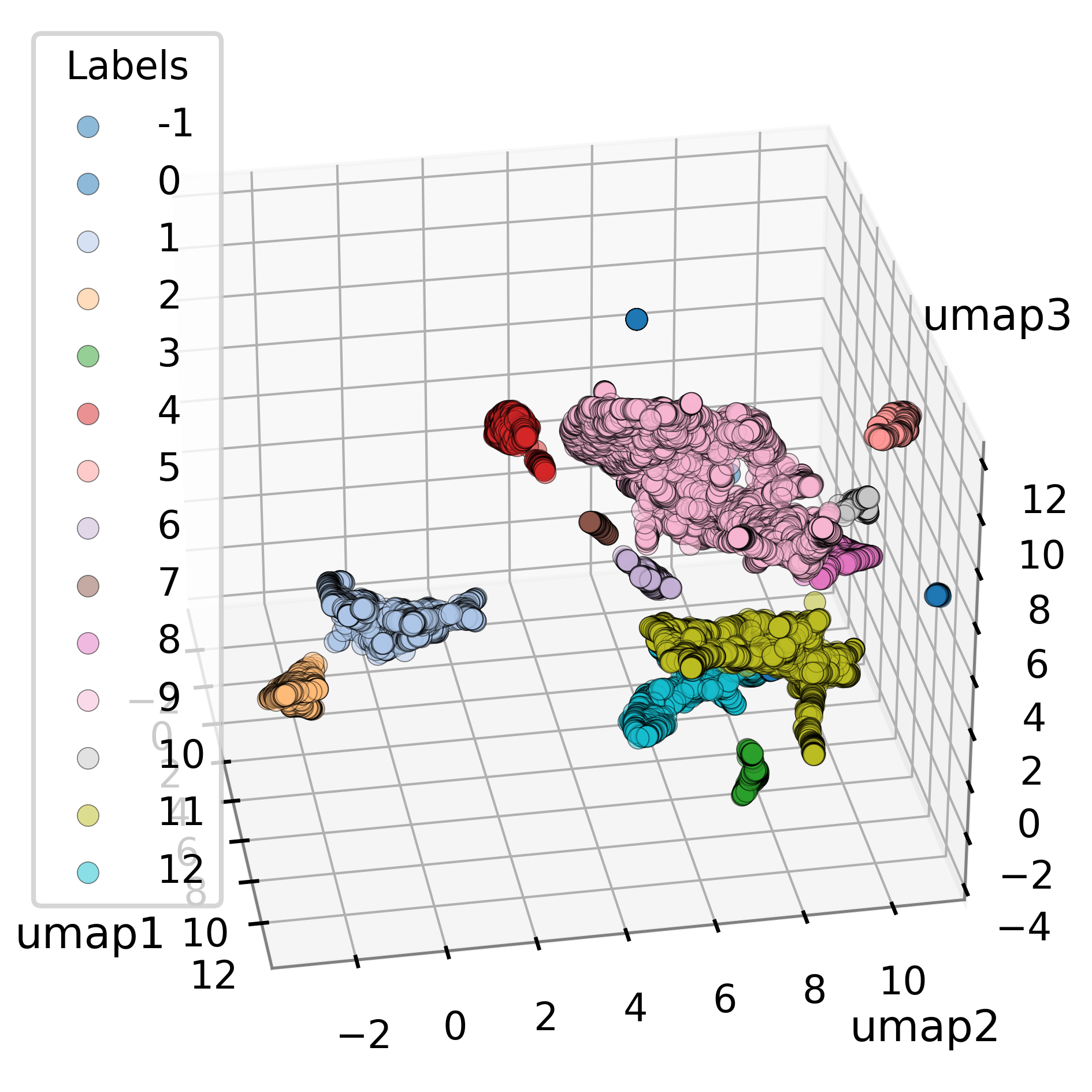} 
    \caption{Coloured clusters for experiment 7, on featureset `mwd' using HDBSCAN for clustering and UMAP for dimension reduction.}
    \label{fig:appendix_mwd_experiment_7_clustercolour} 
\end{figure}

\begin{figure}[!h]
    \centering 
    \includegraphics[width=1.0\textwidth,keepaspectratio]{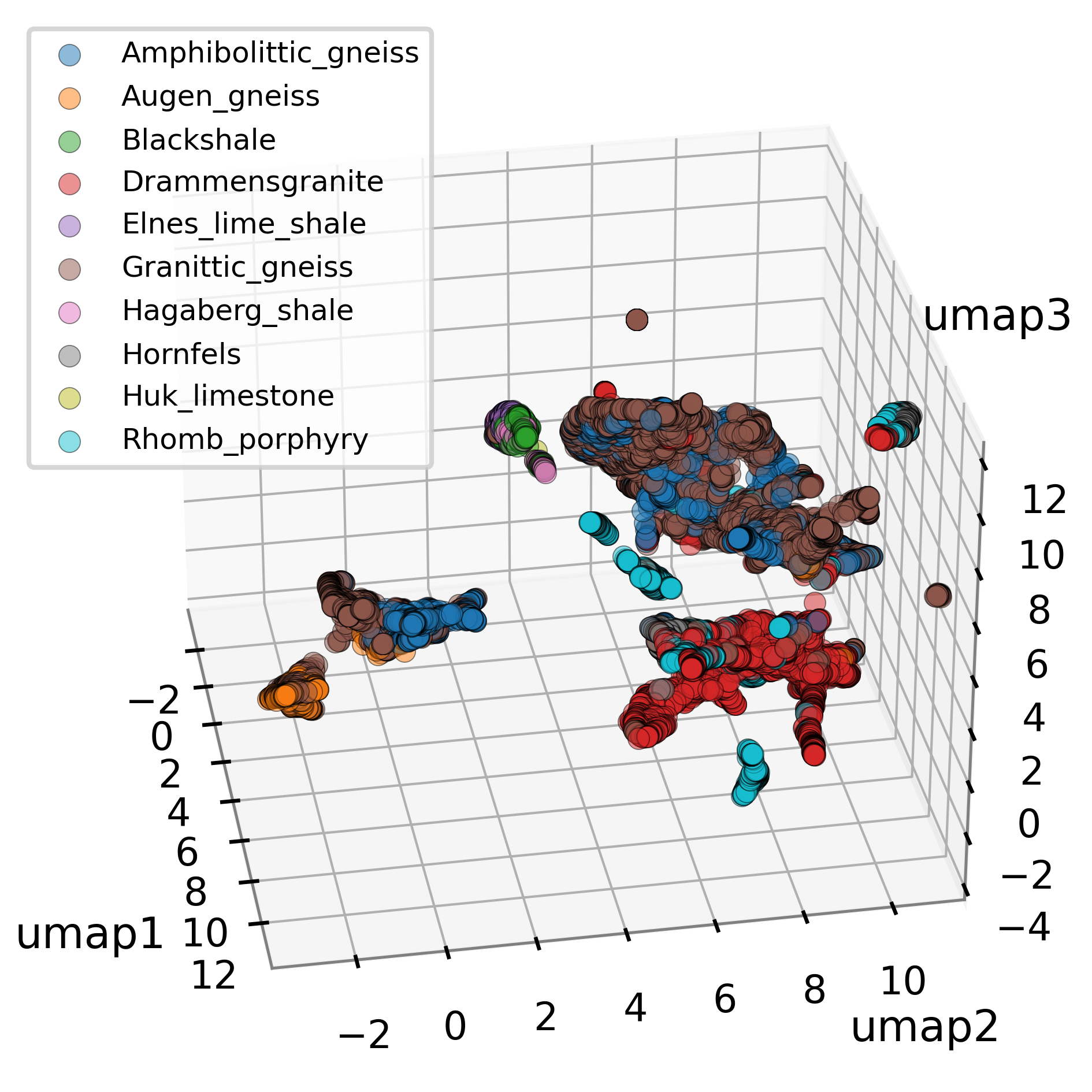} 
    \caption{Clusters labelled with detailed lithology for experiment 7, on featureset `mwd' using HDBSCAN for clustering and UMAP for dimension reduction.}
    \label{fig:appendix_mwd_experiment_7_detailed_geology} 
\end{figure}

\begin{figure}[!h]
    \centering 
    \includegraphics[width=1.0\textwidth,keepaspectratio]{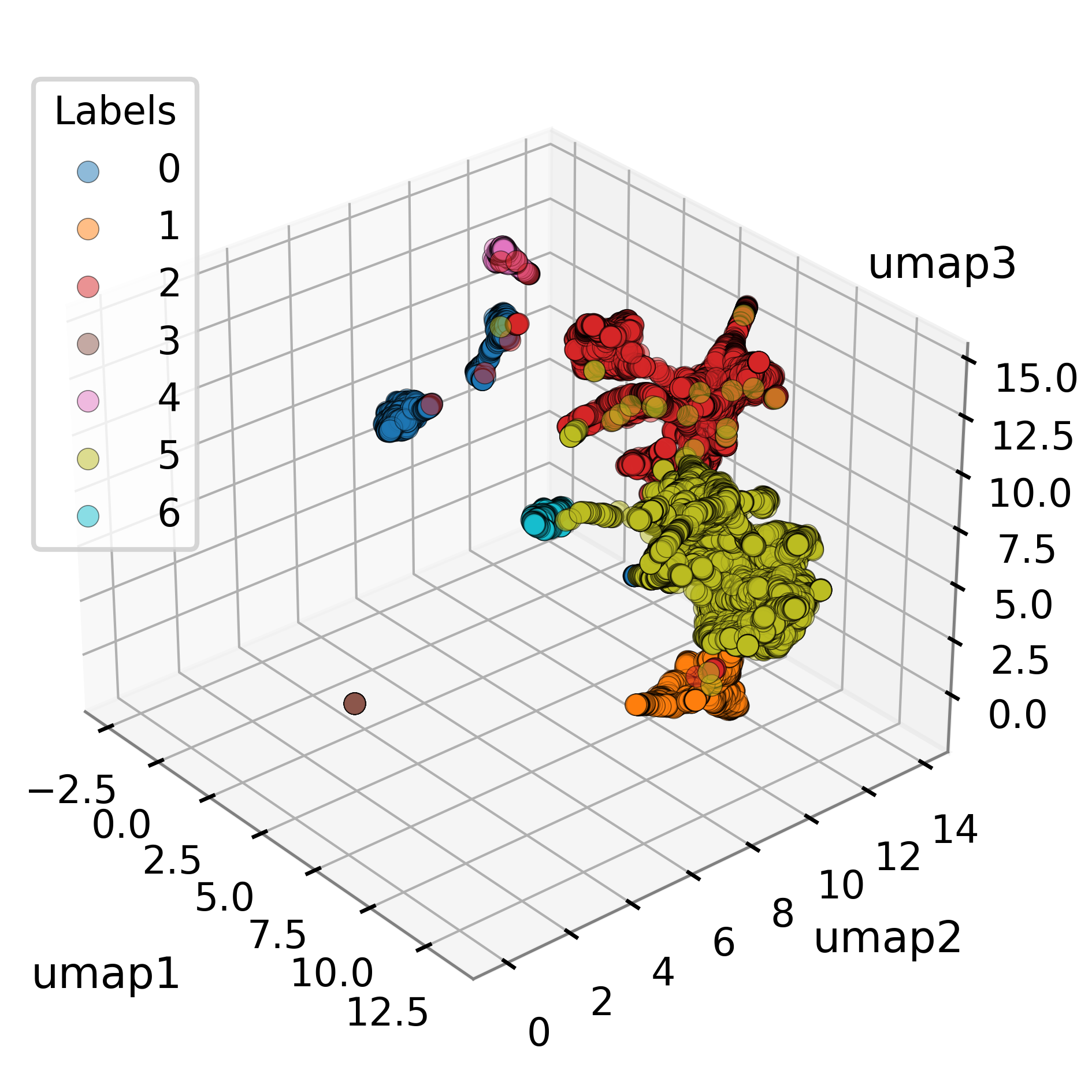} 
    \caption{Colored clusters on featureset `mwd' using Agglomerative Clustering and UMAP for dimension reduction.} 
    \label{fig:mwd_3D_agglomerative} 
\end{figure}

\end{document}